\RequirePackage{fix-cm}
\RequirePackage{amsmath}

\documentclass[twocolumn]{svjour3}          
\smartqed  
\usepackage{graphicx}
\usepackage{hyperref}
\usepackage[square]{natbib} 
\usepackage{comment}
\usepackage{subcaption} 

\newcommand{\eg}{e.g.~}
\newcommand{\ie}{i.e.~}
\newcommand{\etc}{etc.\@}
\usepackage{tikz}
\usepackage{siunitx}
\usepackage{url}

\newcommand{\figLabel}{Fig. }
\newcommand{\secLabel}{Sec. }
\newcommand{\simname}{Sim4CV}

\definecolor{mycolor1}{RGB}{230,0,0}
\definecolor{mycolor2}{RGB}{217,95,2}
\definecolor{mycolor3}{RGB}{117,112,179}
\definecolor{mycolor4}{RGB}{231,41,138}
\definecolor{mycolor5}{RGB}{102,166,30}
\definecolor{mycolor6}{RGB}{230,171,2}
\definecolor{mycolor7}{RGB}{166,118,29}
\definecolor{mycolor8}{RGB}{102,102,102}
\newcommand{\mycirc}[1]{\tikz\draw[{#1},fill={#1}] (0,0) circle (0.8ex);}
\begin{document}

\title{\simname
}
\subtitle{A Photo-Realistic Simulator for Computer Vision Applications}


\author {Matthias M\"uller \and Vincent Casser \and Jean Lahoud \and Neil Smith \and Bernard Ghanem}


\institute{M. M\"uller \and V. Casser \and J. Lahoud \and N. Smith \and B. Ghanem \at
              Electrical Engineering, Visual Computing Center, King Abdullah University of Science and Technology (KAUST), Thuwal, Saudi Arabia \\
              \email{matthias.mueller.2@kaust.edu.sa, \\vincent.casser@gmail.com, jean.lahoud@kaust.edu.sa, neil.smith@kaust.edu.sa, bernard.ghanem@kaust.edu.sa}
}

\date{Received: 18 July 2017 / Accepted: 26 February 2018}

\maketitle

\begin{abstract}
We present a photo-realistic training and evaluation simulator (\simname)\footnote{www.sim4cv.org} with extensive applications across various fields of computer vision. Built on top of the Unreal Engine, the simulator integrates full featured physics based cars, unmanned aerial vehicles (UAVs), and animated human actors in diverse urban and suburban 3D environments. We demonstrate the versatility of the simulator with two case studies: autonomous UAV-based tracking of moving objects and autonomous driving using supervised learning. The simulator fully integrates both several state-of-the-art tracking algorithms with a benchmark evaluation tool and a deep neural network (DNN) architecture for training vehicles to drive autonomously. It generates synthetic photo-realistic datasets with automatic ground truth annotations to easily extend existing real-world datasets and provides extensive synthetic data variety through its ability to reconfigure synthetic worlds on the fly using an automatic world generation tool.


\end{abstract}

\section{Introduction} \label{sec: intro}

\begin{figure*}[ht]
  \centering
  \setlength\tabcolsep{2pt}
  \makebox[\linewidth]{
  \begin{tabular}{cccc}
    Object Tracking
    &Pose Estimation
    &Object Detection
    &Action Recognition \\
    \vspace{-2pt}
    \includegraphics[width=0.245\linewidth,height=0.14\linewidth]{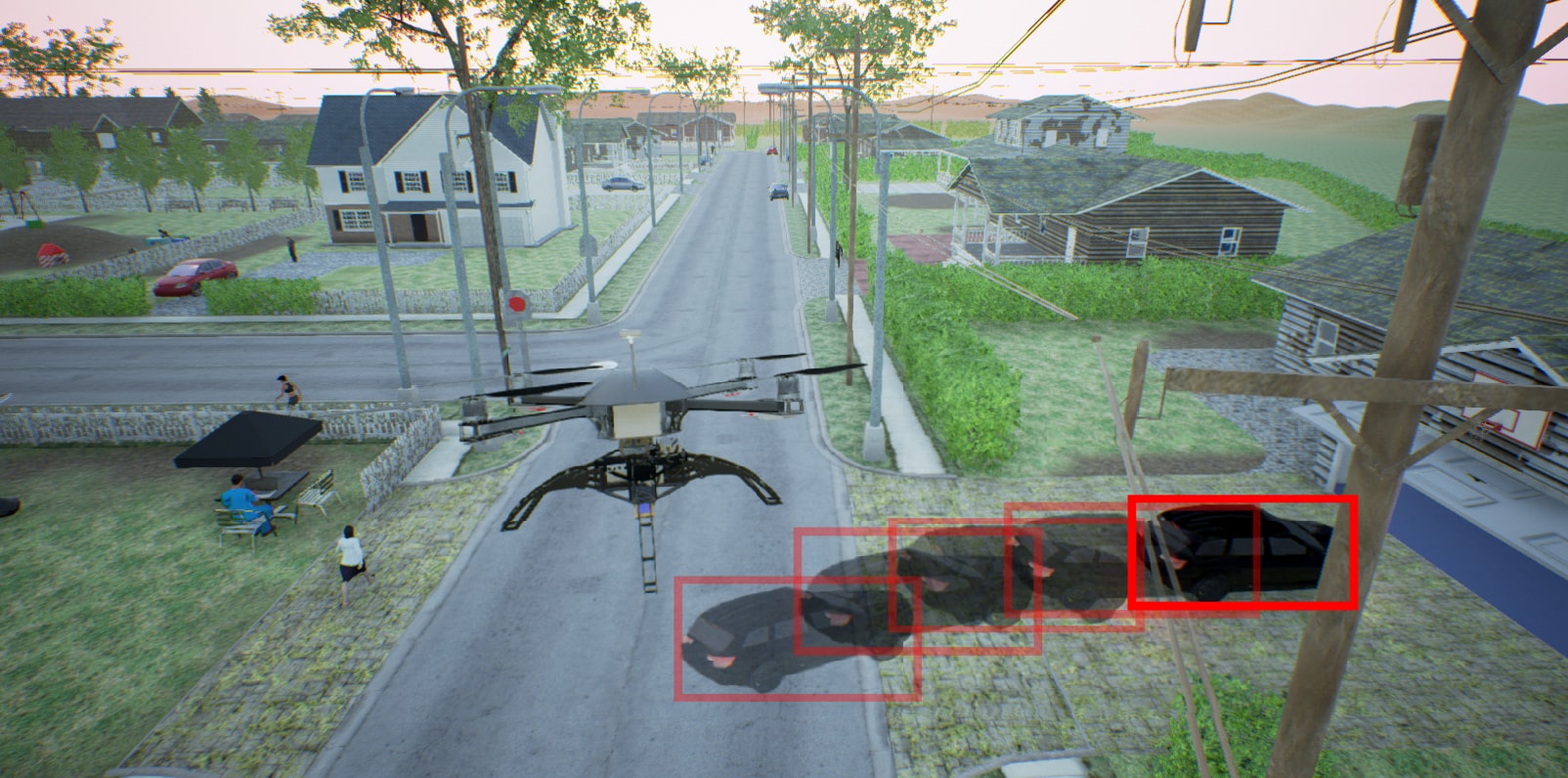}
    & \includegraphics[width=0.245\linewidth,height=0.14\linewidth]{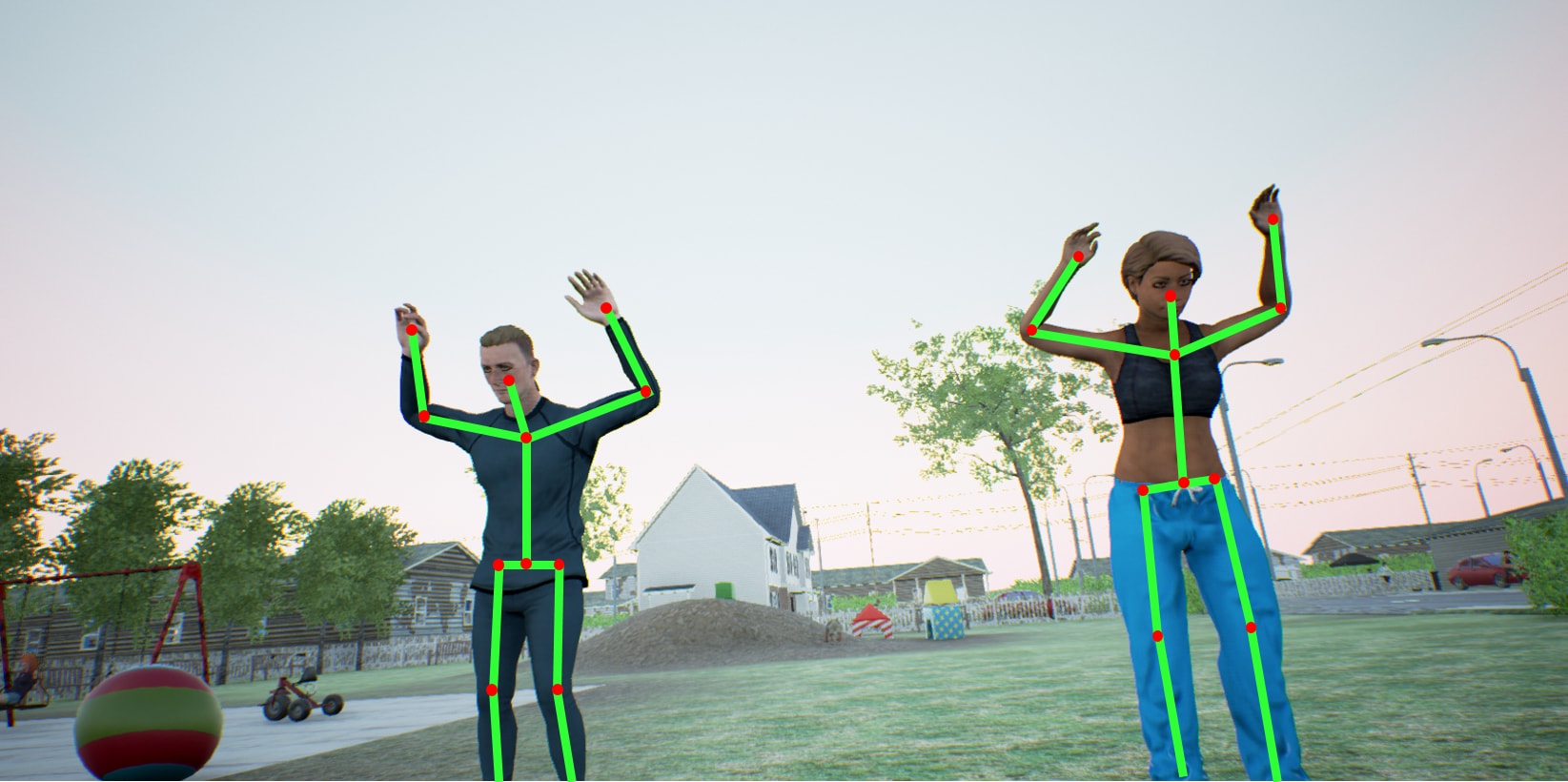}
     & \includegraphics[width=0.245\linewidth,height=0.14\linewidth]{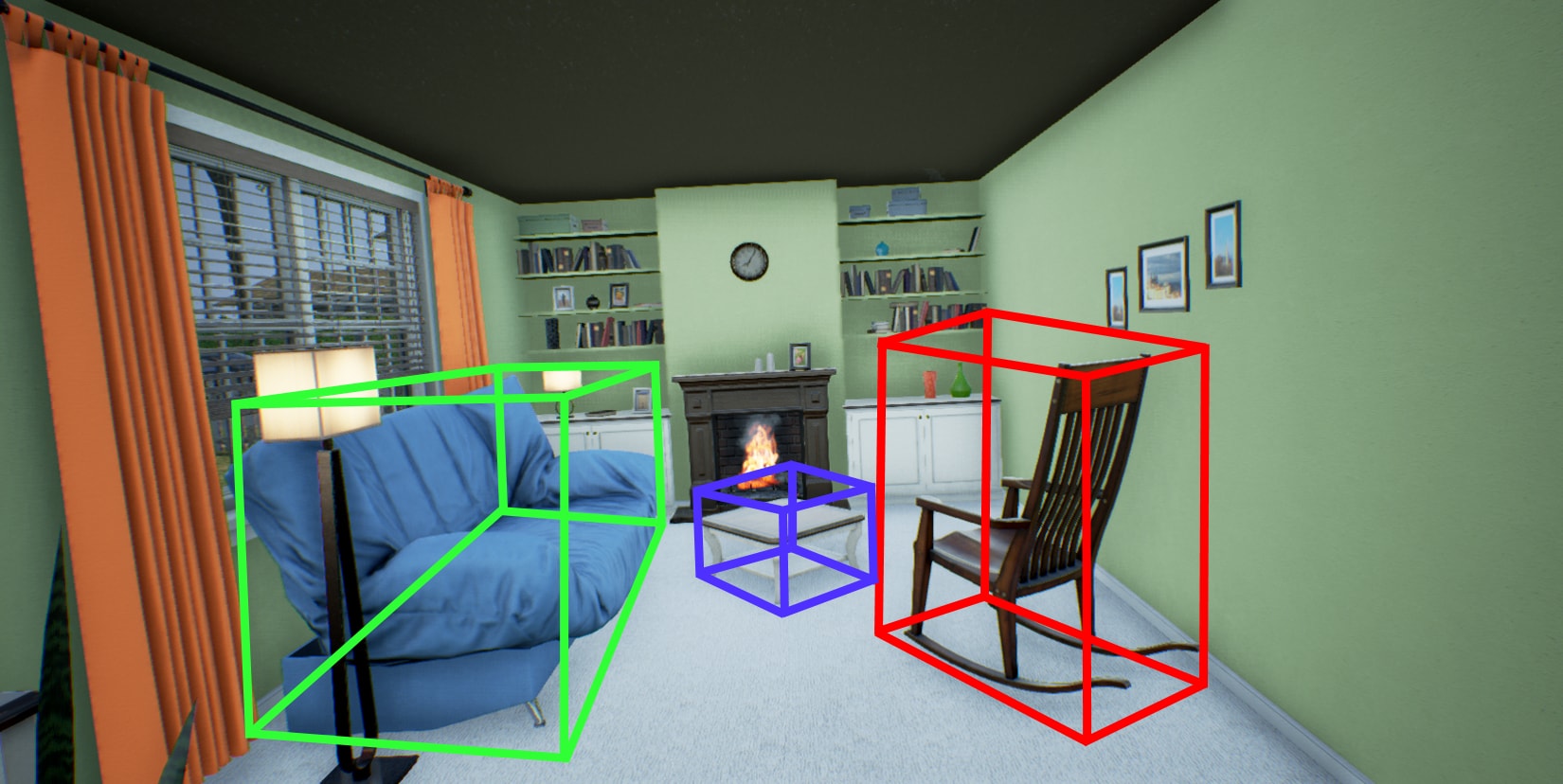}
     & \includegraphics[width=0.245\linewidth,height=0.14\linewidth]{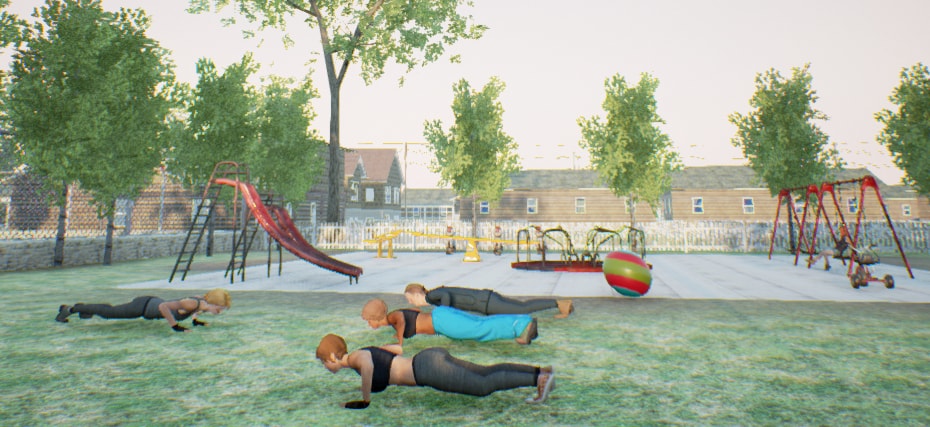} \\
     \mycirc{mycolor1} \mycirc{mycolor3} \mycirc{mycolor4} / \mycirc{mycolor2} 
     & \mycirc{mycolor1} \mycirc{mycolor4} / \mycirc{mycolor2} \mycirc{mycolor3} \mycirc{mycolor7} 
     & \mycirc{mycolor1} \mycirc{mycolor4} / \mycirc{mycolor2} \mycirc{mycolor3} 
     & \mycirc{mycolor1} \mycirc{mycolor3} \mycirc{mycolor5} / \mycirc{mycolor2} \mycirc{mycolor4} 
     \vspace{5pt} \\ 
     Autonomous Navigation
     &3D Reconstruction
     &Crowd Understanding
     &Urban Scene Understanding \\
     \vspace{-2pt}
     \includegraphics[width=0.245\linewidth,height=0.14\linewidth]{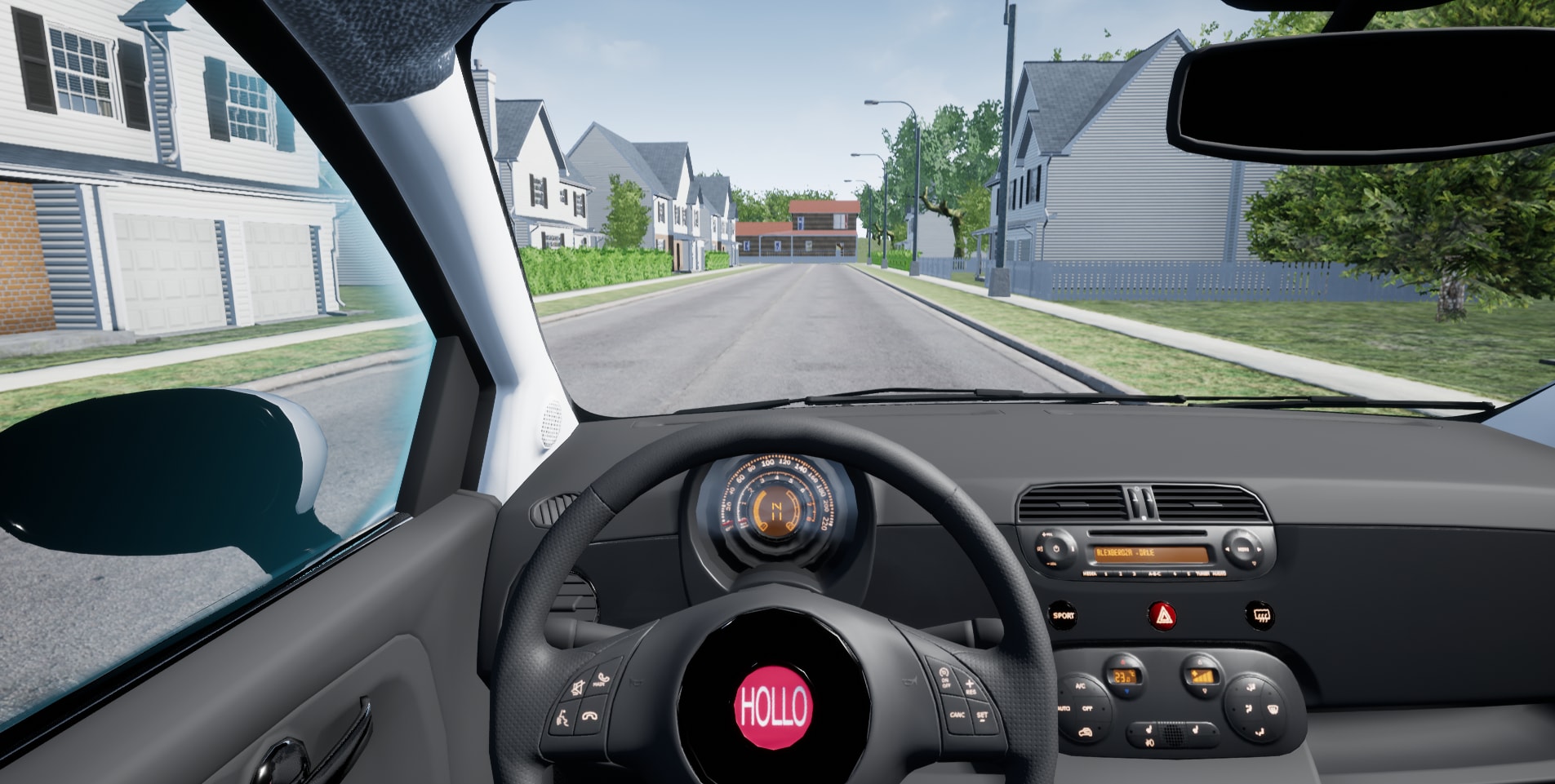}
     & \includegraphics[width=0.245\linewidth,height=0.14\linewidth]{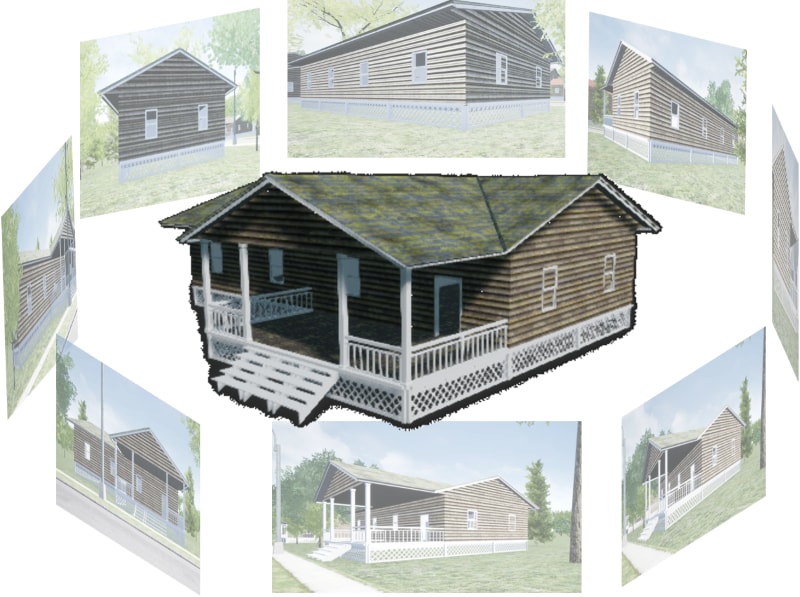}
     & \includegraphics[width=0.245\linewidth,height=0.14\linewidth]{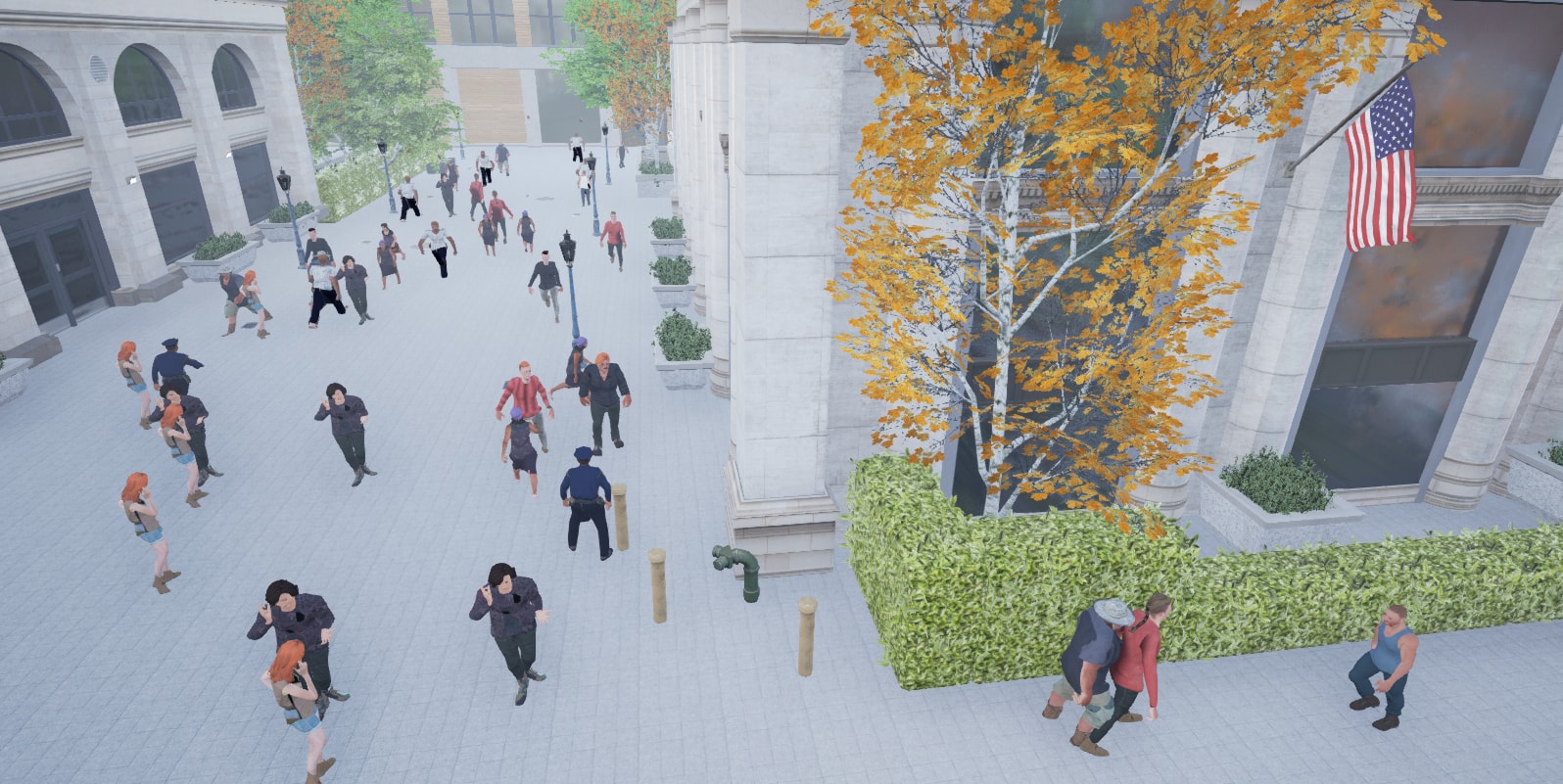}
     & \includegraphics[width=0.245\linewidth,height=0.14\linewidth]{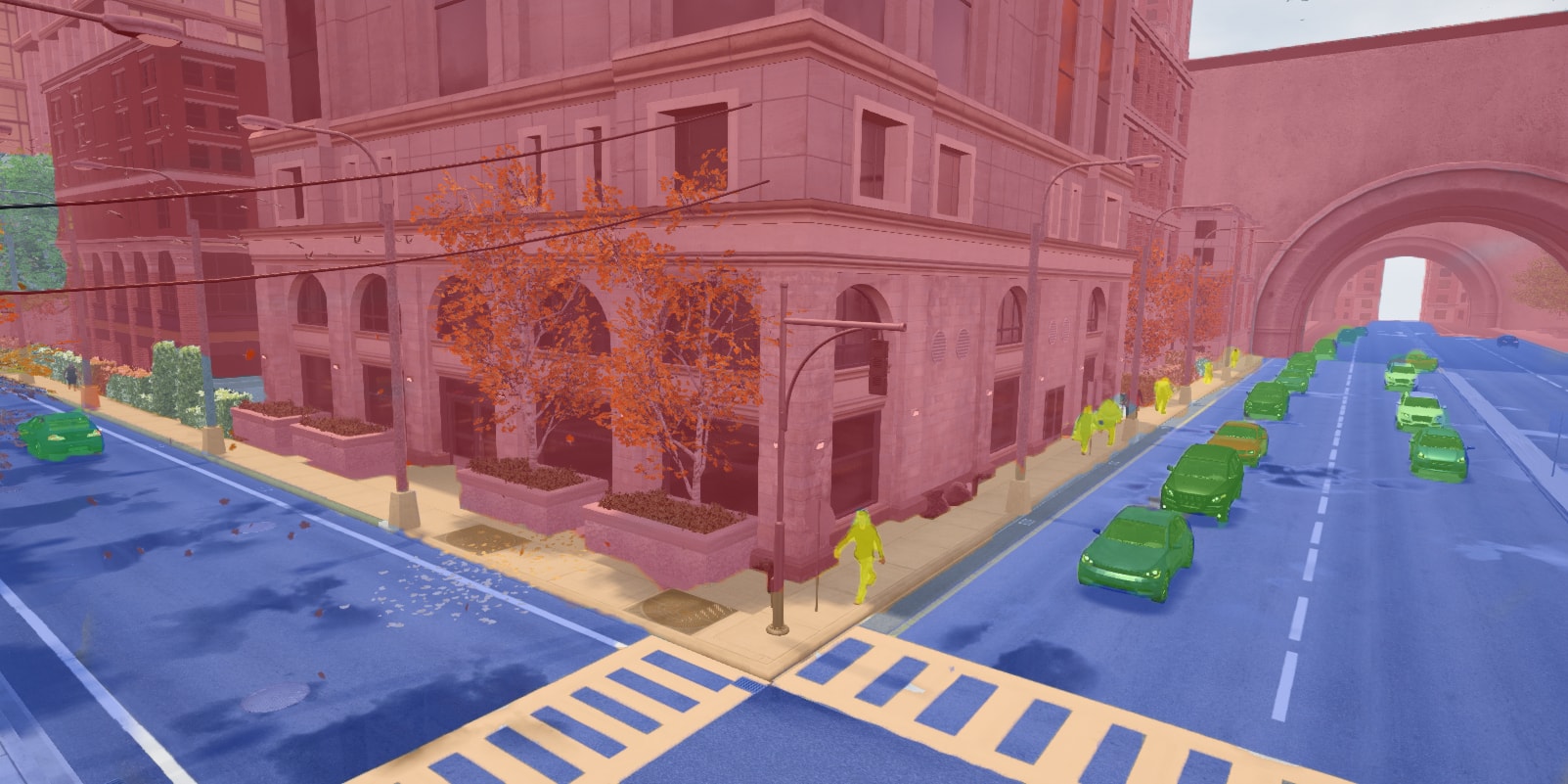} \\
     \mycirc{mycolor1} \mycirc{mycolor3} \mycirc{mycolor5} / \mycirc{mycolor2} \mycirc{mycolor7} 
     & \mycirc{mycolor1} \mycirc{mycolor2} \mycirc{mycolor8} / \mycirc{mycolor3} 
     & \mycirc{mycolor1} \mycirc{mycolor4} / \mycirc{mycolor2} \mycirc{mycolor3} \mycirc{mycolor5} 
     & \mycirc{mycolor1} \mycirc{mycolor4} / \mycirc{mycolor2} \mycirc{mycolor3} 
     \vspace{5pt} \\
     Indoor Scene Understanding
     &Multi-agent Collaboration
     &Human Training
     &Aerial Surveying \\ 
     \vspace{-2pt}
     \includegraphics[width=0.245\linewidth,height=0.14\linewidth]{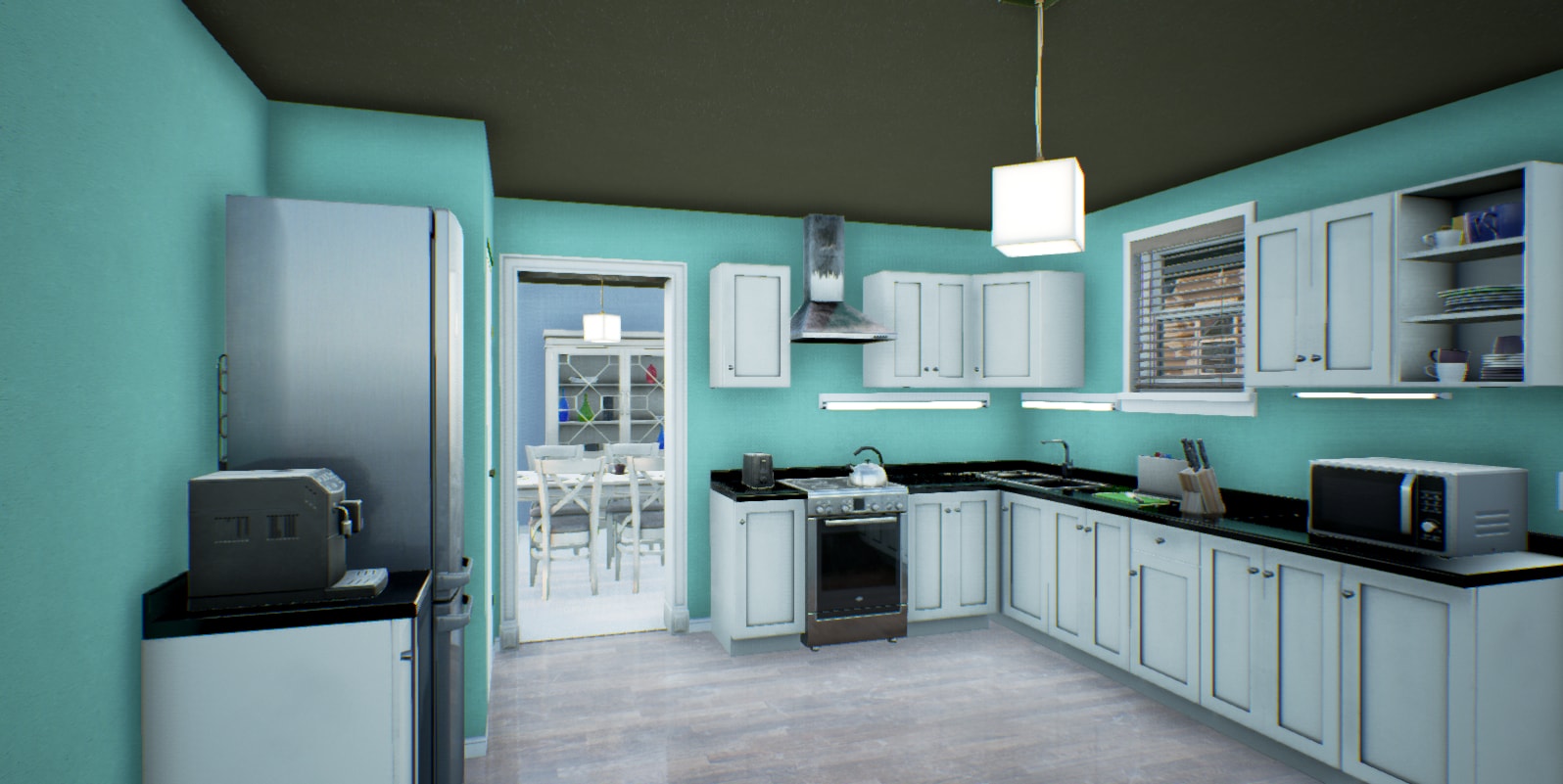}
     & \includegraphics[width=0.245\linewidth,height=0.14\linewidth]{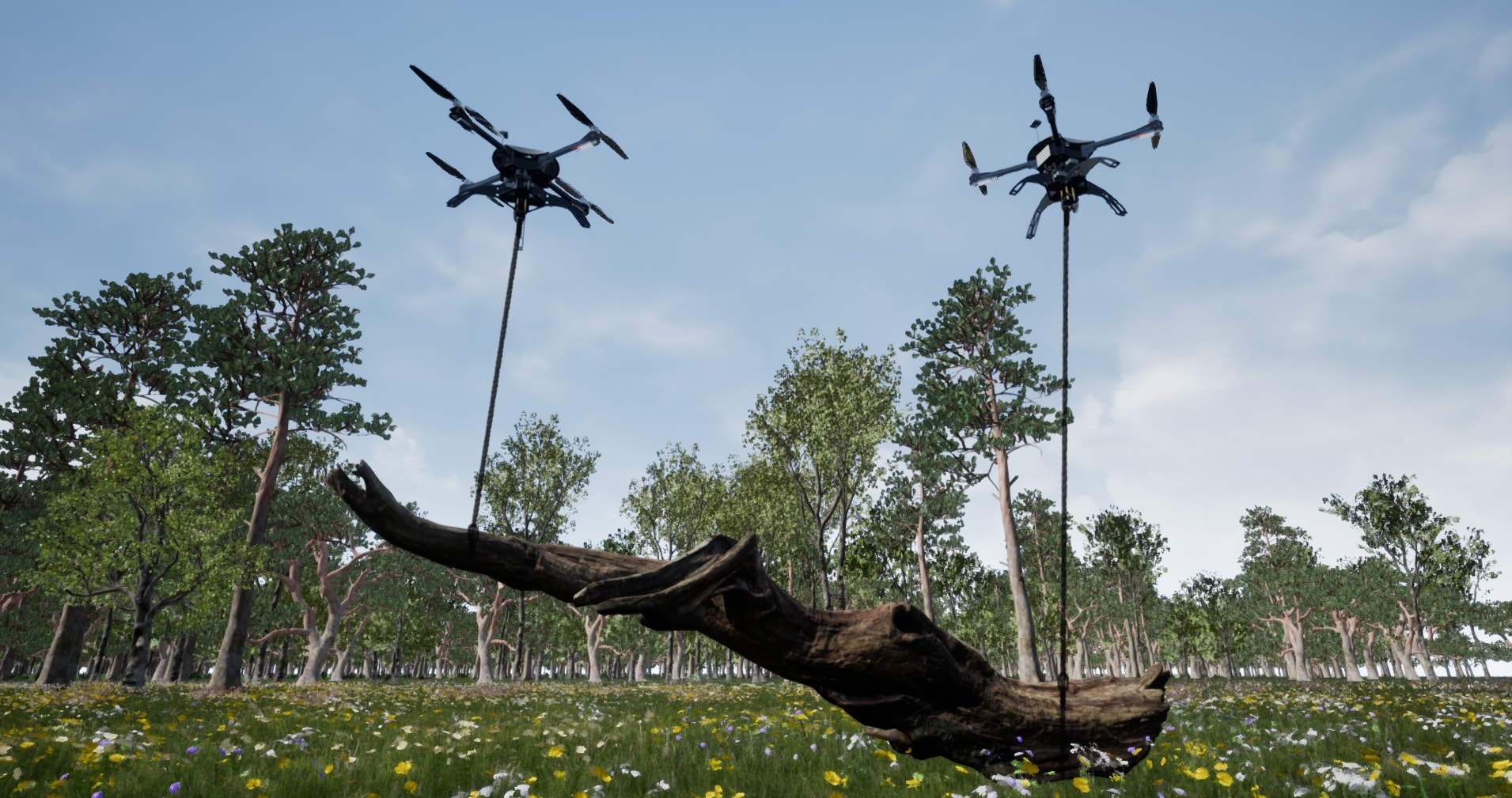}
     & \includegraphics[width=0.245\linewidth,height=0.14\linewidth]{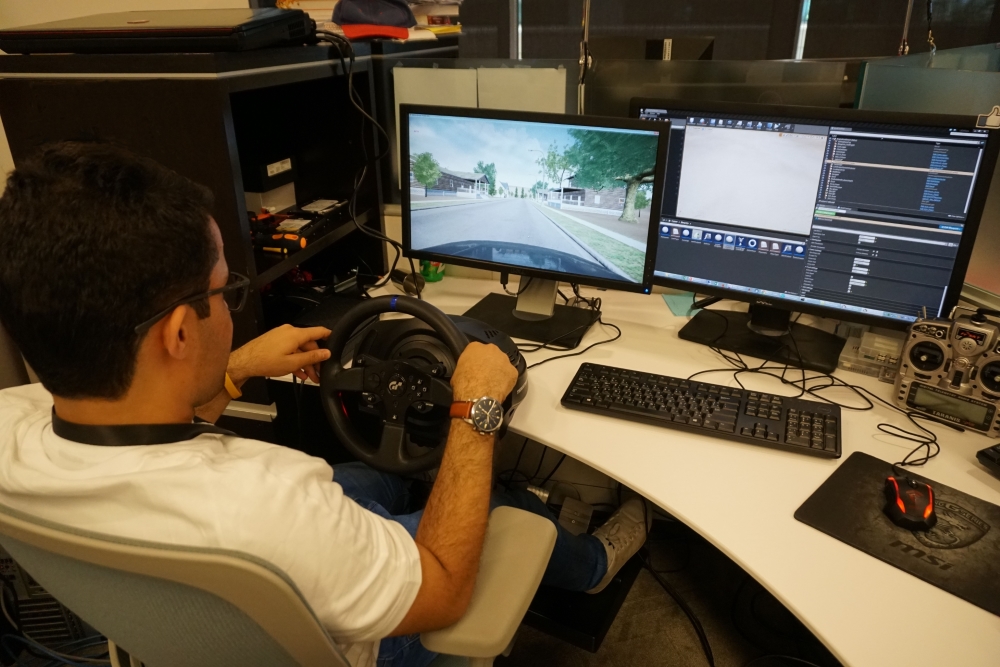}
     & \includegraphics[width=0.245\linewidth,height=0.14\linewidth]{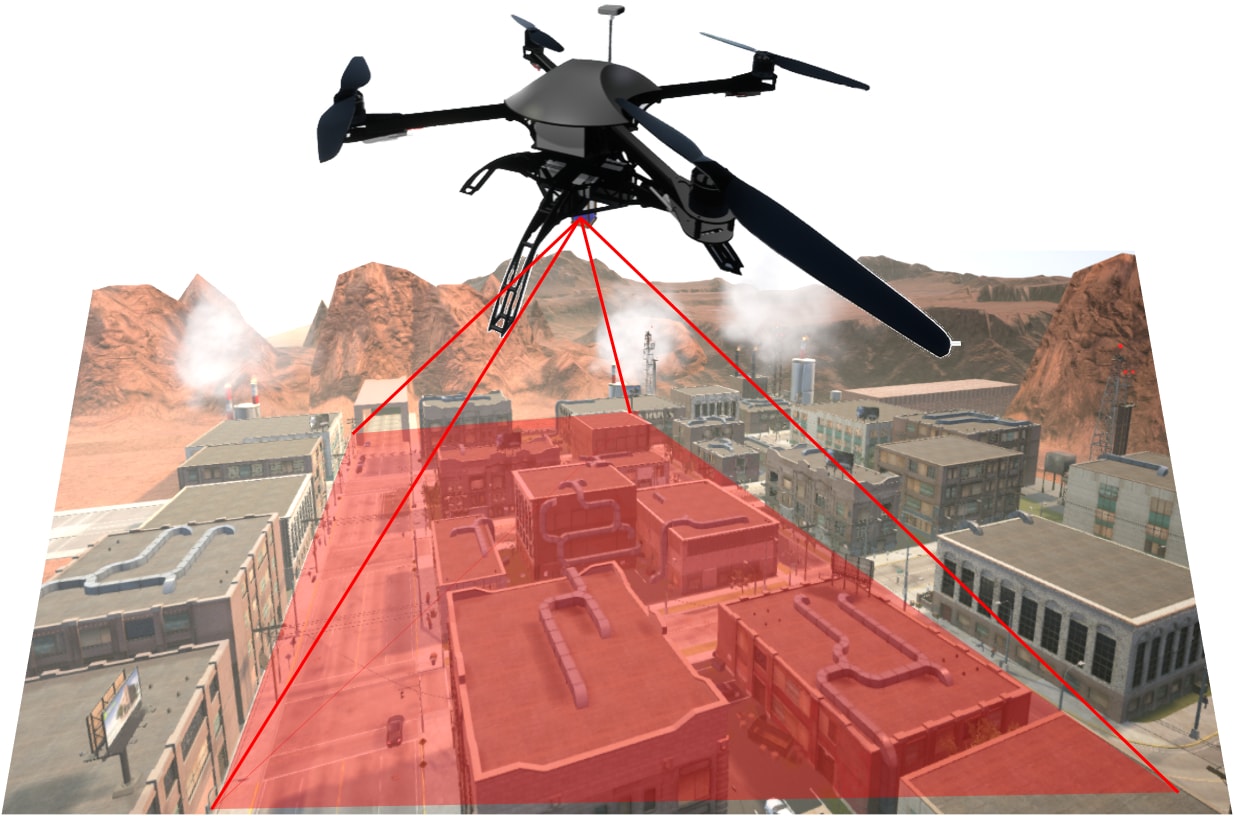} \\
     \mycirc{mycolor1} \mycirc{mycolor2} \mycirc{mycolor4} / \mycirc{mycolor3} 
     & \mycirc{mycolor1} \mycirc{mycolor3} / \mycirc{mycolor2} \mycirc{mycolor4} \mycirc{mycolor7} \mycirc{mycolor8} 
     & \mycirc{mycolor1} \mycirc{mycolor3} \mycirc{mycolor6} \mycirc{mycolor7} 
     & \mycirc{mycolor1} \mycirc{mycolor2} \mycirc{mycolor8} / \mycirc{mycolor3} \mycirc{mycolor4}  
     \vspace{5pt} \\
     \multicolumn{4}{c}{
     \makebox[3.5cm][l]{\tikz\draw[mycolor1,fill=mycolor1] (0,0) circle (0.8ex); 
     Image}
     \makebox[3.5cm][l]{\tikz\draw[mycolor2,fill=mycolor2] (0,0) circle (0.8ex); 
     Depth/Multi-View}
     \makebox[3cm][l]{\tikz\draw[mycolor3,fill=mycolor3] (0,0) circle (0.8ex); 
     Video}
     \makebox[3.5cm][l]{\tikz\draw[mycolor4,fill=mycolor4] (0,0) circle (0.8ex); 
     Segmentation/Bounding Box}}\\
     \multicolumn{4}{c}{
     \makebox[3.5cm][l]{\tikz\draw[mycolor5,fill=mycolor5] (0,0) circle (0.8ex); 
     Image Label}
     \makebox[3.5cm][l]{\tikz\draw[mycolor6,fill=mycolor6] (0,0) circle (0.8ex); 
     User Input}
     \makebox[3cm][l]{\tikz\draw[mycolor7,fill=mycolor7] (0,0) circle (0.8ex); 
     Physics}
     \makebox[3.5cm][l]{\tikz\draw[mycolor8,fill=mycolor8] (0,0) circle (0.8ex); 
     Camera Localization}
     }
   \end{tabular}
   }
   \caption{An overview of some common applications in computer vision in which our simulator can be used for generating synthetic data and performing (real-time) evaluation. For each application, we show which type of data is usually a necessity (\eg segmentation masks/bounding box annotations for learning object detection) and which can be used optionally.}\label{figtab}
 \end{figure*}

The photo-realism of modern game engines provides a new avenue for developing and evaluating methods for diverse sets of computer vision (CV) problems, which will ultimately operate in the real-world. In particular, game engines such as U\-ni\-ty, U\-NI\-GINE, CRY\-EN\-GINE, and Unreal Engine 4 (UE4) have begun to open-source their engines to allow low-level redesign of core functionality in their native programming languages. This full control over the gaming engine allows researchers to develop novel applications and lifelike physics based simulations, while benefiting from the free generation of photo-realistic synthetic visual data. Since these modern game engines run in real-time, they provide an end-to-end solution for training with synthetic data, conducting controlled experiments, and real-time benchmarking. As they approach not only photo-realism but lifelike physics simulation, the gap between simulated and real-world applications is expected to substantially decrease. In this paper, we present \simname, a full featured, customizable, physics-based simulator built with\-in the Unreal Engine 4. The simulator directly provides accurate car and UAV physics, as well as, both the latest state-of-the-art tracking algorithms with a benchmark evaluation tool and a Ten\-sor\-Flow-based deep learning interface.

\simname~allows access to both visual data captured from cameras mounted in the simulated environment and semantic information that can be used for learning-based CV applications. For example, in addition to RGB images, the simulator provides numerous capabilities, such as depth, segmentation, and ground truth labelling, which can enable a wide variety of applications as shown in \figLabel \ref{figtab}. More details on the simulator capabilities and applications are presented in \secLabel \ref{sec: simulator}. Although recent work by \cite{GtaV} has shown the advantages of using pre-built simulated worlds (\eg GTA V's Los Angeles city), this approach is not easily re-configurable. While these worlds are highly detailed, they are not amenable to user customization, which limits the potential variety needed for large-scale data generation and extensive evaluation in diverse scenarios. To address this drawback and unlike other simulators used for CV purposes, we divide a large variety of high-poly Physically-Based Rendering (PBR) textured assets into building blocks that can be placed and configured within a simple GUI and then procedurally generated within \simname~at runtime. Moreover, dynamic agents (\eg pedestrians and cars) can be included to generate more dynamic scenarios. Of course, the open source nature of the implementation gives the user the freedom to modify/prune/enrich this set of assets. A similar strategy can be taken for indoor scenes as well.  As such, this process can generate a very rich variety of city and suburban scenes, thus, bolstering the generation of diverse datasets for  deep neural network (DNN) training, as a step towards preventing the over-fitting of DNN methods and fostering better generalization properties. To advocate the generality of \simname, we adopt two popular use cases from the CV literature: real-time tracking evaluation from a UAV and autonomous car driving. 

Empowering UAVs with automated CV capabilities (\eg tracking, object/activity recognition, mapping, \etc) is becoming a very important research direction in the field and is rapidly accelerating with the increasing availability of low-cost, commercially available UAVs. In fact, aerial tracking has enabled many new applications in computer vision (beyond those related to surveillance), including  search and rescue, wild-life monitoring, crowd monitoring/management, navigation/localization, obstacle/object avoidance, and videography of extreme sports. Aerial tracking can be applied to a  diverse set of objects (\eg humans, animals, cars, boats), many of which cannot be physically or persistently tracked from the ground. In particular, real-world aerial tracking scenarios pose new challenges to the tracking problem, exposing areas for further research. In \simname, one can  directly feed video frames captured from a camera onboard a UAV  to CV trackers and retrieve their tracking results to update UAV flight. Any tracker (\eg written in MATLAB, Python, or C++) can be tested within the simulator across a diverse set of photo-realistic simulated scenarios allowing new quantitative methods for evaluating tracker performance. In fact, this paper extends our previous work \cite{Mueller2016} by evaluating trackers on a new more extensive synthetic dataset. 

Inspired by recent work on self-driving cars by \cite{NvidiaCar} and the synthetic KITTI dataset by \cite{gaidon2016virtual}, we implement a TensorFlow-based DNN  interface with \simname. In our self-driving application, both \simname~and the DNN  run at real-time speeds in parallel, allowing fully interactive evaluation to be performed. The DNN is trained in a supervised manner by exploiting  the extensive and diverse synthetic visual data that can be captured and the free accompanying labelling (\ie waypoint coordinates). With some required acceleration measures, our simulator can also be extended to facilitate reinforcement learning methods for this task, although this remains outside the scope of this paper. 

\paragraph{Contributions.} The contributions of our work are four-fold: \textbf{(1)} An end-to-end physics-based, fully customizable, open-source simulator environment for the CV community working on autonomous navigation, tracking, and a wide variety of other applications; \textbf{(2)} a customizable synthetic world generation system; \textbf{(3)} a novel approach for tracker evaluation with a high-fidelity real-time visual tracking simulator; and \textbf{(4)} a novel, robust deep learning based approach for autonomous driving that is flexible and does not require manually collected training data.

\section{Related Work} \label{sec: related work}

\subsection{Learning from Simulation} A broad range of work has recently exploited physics based simulators for learning purposes, namely in animation and motion planning (\cite{birdFlightSimulator,unity3Dphysics,deepReinforcementSimulator,UE4simulator,bikeStunts,balancingSimulator,parkourSimulation}), scene understanding (\cite{Battaglia05112013,syntheticRGBD}), pedestrian detection (\cite{Pedestrian2010}), and identification of 2D/3D objects (\cite{syntheticCarRecognition,syntheticVehicleTraining,teaching3D}). For example, in \cite{birdFlightSimulator}, a physics-based computer game environment (Unity) is used to teach a simulated bird to fly. Moreover, hard\-ware-in-the-loop (HIL) simulation has also been used in robotics to develop and evaluate controllers and for visual servoing studies (\eg JMAVSim, \cite{uavHIL2015,hilUAV} and RotorS by \cite{Frrr2016}). The visual rendering in these simulators is often primitive and relies on off-the-shelf simulators (\eg Realflight, Flightgear or XPlane). They do not support advanced shading and post-processing techniques, are limited in terms of available assets and textures, and do not support motion capture (MOCAP) or key-frame type animation to simulate natural movement of actors or vehicles.

Recent work (\eg \cite{gaidon2016virtual,GtaV,de2016procedural, RosCVPR16, airsim2017fsr, carla}) show the advantages of exploiting the photo-realism of modern game engines to generate training datasets and pixel-accurate segmentation masks. Since one of the applications presented in this work is UAV tracking, we base our own \simname~simulator on our recent work  (\cite{Mueller2016}), which provides full hardware and software in-the-loop UAV simulation built on top of the open source Unreal Engine 4. Following this initial work, plugins have been created that enable specific features that are not present in Unreal Engine 4, such as physics simulation and  annotation generation (\eg segmentation masks). For example, \cite{airsim2017fsr} developed AirSim, a plugin that simulates the physics and control of a UAV from a flight controller, but leaves the development of game design and integration with external applications to the user. The physics of the UAV is evaluated outside UE4, preventing full exploitation and interaction within the rich dynamic physics environment. This limits UAV physics to simple collision events and simulated gravity. Similarly, UnrealCV by \cite{qiu2017unrealcv} provides a plugin with a socket-based communication protocol to interact with Matlab/Python code providing generic text and image based command and response communication. The authors also provide a tutorial on how to combine it with OpenAI Gym (\cite{openaigym}) for training and evaluating visual reinforcement learning. 
Nevertheless, in order to utilize UnrealCV for a specific vision application, extensive work is still needed both in the vision program and UE4 (\eg UE4 world content creation and setup).

In contrast, \simname~is a fully integrated tool that does not require extensive game development in UE4 or a host vision application to conduct a simulation experiment. It provides a complete integrated system for supervised and reinforcement learning based approaches for driving and flying, as well as, real-time object tracking with visual servoing in a dynamic and changeable world. It provides a complete integrated system that combines 3 main features into one package: \textbf{(1)} automatic world generation that enables easy and fast creation of diverse environments, \textbf{(2)} communication interface that can be used with Matlab, C++, and Python, and \textbf{(3)} fully implemented \simname~applications for UAV-based tracking and autonomous driving, which researchers can immediately exploit to build better algorithms for these purposes. All these features are enriched with rich content and high versatility, enabling straightforward integration of other tools into \simname. 

\subsection{UAV Tracking}
A review of related work indicates that there is still a limited availability of annotated datasets specific to UAVs, in which trackers can be rigorously evaluated for precision and robustness in airborne scenarios. Existing annotated video datasets include very few aerial sequences (\cite{28}). Surveillance datasets such as PETS or CAVIAR focus on static surveillance and are outdated. VIVID \cite{VIVID} is the only publicly available dedicated aerial dataset, but it is outdated and has many limitations due to its small size (9 sequences), very similar and low-resolution sequences (only vehicles as targets), sparse annotation (only every 10th frame), and focus on higher altitude, less dynamic fixed-wing UAVs. There are several recent benchmarks that were created to address specific deficiencies of older benchmarks and introduce new evaluation approaches \cite{nuspro,ColorBenchmark,VisualTrackingSurvey}, but they do not introduce videos with many tracking nuisances addressed in this paper and common to aerial scenarios.

Despite the lack of benchmarks that adequately address aerial tracking, the development of tracking algorithms for UAVs has become very popular in recent years. The majority of  object tracking methods employed on UAVs rely on feature point detection/tracking (\cite{3,Nussberger2014}) or color-centric object tracking  (\cite{Kendall2014}). Only a few works in the literature (\cite{uavtld}) exploit more accurate trackers that commonly appear in generic tracking benchmarks such as MIL in \cite{18,uavmil}, TLD in \cite{uavtld}, and STRUCK in  \cite{icraMAVtracker,mueller_iros16}. There are also more specialized trackers tailored to address specific problems and unique camera systems such as in wide aerial video (\cite{Pollard2012,Prokaj2014}), thermal and IR video  (\cite{14,Portmann2014}), and RGB-D video  (\cite{naseer13iros}). In \simname, the aforementioned trackers, as well as, any other tracker, can be integrated into the simulator for real-time aerial tracking evaluation, thus, standardizing the way trackers are compared in a diverse and dynamic setting. In this way, state-of-the-art trackers can be extensively tested in a life-like scenario before they are deployed in the real-world.

\subsection{Autonomous Driving}
Work by \cite{NvidiaCar,deepDriving,ForestTrail,pomerleau1989alvinn,NIPS2005,Andersson2017,Kim2015DeepNN,Shah:2016} show that with sufficient training data and augmented camera views autonomous driving and flight can be learned by a DNN. The driving case study for \simname~is primarily inspired by \cite{deepDriving} and other work (\cite{mnih2016asynchronous, deepReinforcementSimulator,Koutnik:2013,Koutník2014}), which uses TORCS (The Open Racing Car Simulator by \cite{torcs}) to train a DNN to drive at casual speeds through a course and properly pass or follow other vehicles in its lane. The vehicle controls are predicted in \cite{deepDriving} as a discrete set of outputs: turn-left, turn-right, throttle, and brake. The primary limitation of TORCS for DNN development is that the environment in which all training is conducted is a race track with the only diversity being the track layout. This bounded environment does not include the most common scenarios relevant to autonomous driving, namely urban, suburban, and rural environments, which afford complexities that are not present in a simple racing track (\eg pedestrians, intersections, cross-walks, 2-lane on-coming traffic, \etc). 

The work of \cite{GtaV} proposed an approach to extract synthetic visual data directly from the Grand Theft Auto V (GTA V) computer game. Of particular interest is the high-quality, photo-realistic urban environment in which rich driving data can be extracted, along with the logging of user input for supervised (SL) or reinforcement learning (RL). However, a primary limitation of this approach is that the virtual world, although generally dynamic and highly photo-realistic, cannot be interactively controlled or cus\-to\-mized. This limits its flexibility in regards to  data augmentation, evaluation, and repeatability, thus, ul\-ti\-mately affecting the generalization capabilities of DNNs trained in this setup. One insight we highlight in this paper is that without adding additional shifted views, the purely visual driving approach overfits and fails. Since synthetic data from GTA V is limited to a single camera view, there is no possible way to augment the data or increase the possible variety of views. Second, repeatability is not possible either, since one cannot start an evaluation in the exact spot everytime, control the randomized path of cars and people in the world, or programmatically control or reconfigure anything else in the game. 

Virtual KITTI by \cite{gaidon2016virtual} provides a  Unity based simulation environment, in which real-world video sequences can be used as input to create virtual and realistic proxies of the real-world. They demonstrate that the gap is small in transferring between DNN learning on synthetic videos and their real-world counterparts, thus, emphasizing the crucial need for  photo-realistic simulations. They generate 5 cloned worlds and 7 variations of each to create a large dataset of 35 video sequences with about 17,000 frames. The synthetic videos are automatically annotated and serve as ground truth for RGB tracking, depth, optical flow, and scene segmentation. This work is primarily focused on generating video sequences and  does not explore actual in-game mechanics, such as the physics of driving the vehicles, movement of actors within the world, dynamic scene interactions, and real-time in-world evaluation. Control predictions using DNNs is not addressed in their work, since the focus is on the evaluation of trained DNN methods on the video sequences extracted from the engine. Our work addresses these unexplored areas, building upon a fundamental conclusion of \cite{gaidon2016virtual}, which states that proxy virtual worlds do have high transferability to the real-world (specifically for DNN based methods) notably in the case of autonomous driving.

Although \cite{NvidiaCar} collected datasets primarily from the real-world in their DNN autonomous driving work, their approach to create additional synthetic images and evaluate within a simulation is very relevant to the work presented here. The simulated and on-road results of their work demonstrate advances in how a DNN can learn end-to-end the control process of a self-driving car directly from raw input video data. However, the flexibility regarding augmentation of data collected in the real-world is strongly constrained, so much so, that they had to rely on artificial view interpolations to get a limited number of additional perspectives. \simname~does not share this disadvantage, since it can generate any amount of data needed in the simulator, as well as, evaluate how much and what type of view augmentation is needed for the best performing DNN. 
Furthermore, our DNN approach for the \simname~driving case study does not require human training data, and it possesses a number of additional advantages by design, such as flexibility regarding vehicle controllers and easy support for lane changing, obstacle avoidance, and guided driving compared to an end-to-end approach.

\section{Simulator Overview} \label{sec: simulator}
\vspace{3pt}\noindent\textbf{Setup.} \simname~ is built like a video game that can simply be installed without any additional dependencies making it very easy to use. It comes with a full graphical user interface, through which users can modify all relevant settings. It is also possible to modify the underlying configuration files directly or by means of a script. The binaries of our simulator contain rich worlds for driving and flying, several vehicles (two passenger cars, one RC truck and two UAVs), and several carefully designed maps. Our external map editor (see \figLabel \ref{fig:city_editor}) allows users to create their own maps. The communication interface allows external programs to receive images and state information of the vehicle from the simulator and send control signals to the simulator. We provide examples for C++, Python and Matlab. In addition to the packaged simulator for easy use, we also plan to release a developer version with all the source code to allow the community to build on top of our work, use plugins such as AirSim (\cite{airsim2017fsr}) or UnrealCV (\cite{qiu2017unrealcv}) with \simname, and make contributions to our simulator project.

\vspace{3pt}\noindent\textbf{Simulator capabilities.} \simname~is built on top of Epic Game's Unreal Engine 4 and expands upon our previous work in \cite{Mueller2016}, which primarily focused on UAV simulation. As recognized by others (\cite{gaidon2016virtual,GtaV}), modern game engine architecture allows real-time rendering of not only RGB images, but can also with minor effort be re-tasked to output pixel-level segmentation, bounding boxes, class labels, and depth. Multiple cameras can be setup within a scene, attached to actors, and programmatically moved at each rendering frame. This capability allows  additional synthetic data to be generated, such as simultaneous multi-view rendering, stereoscopy, structure-from-motion, and view augmentation. UE4 also has an advanced physics engine allowing the design and measurement of complex vehicle movement. Not only does the physics allow realistic simulation of moving objects but it can also be coupled with  additional physics measurements at each frame. Finally, the broad support of flight joysticks, racing wheels, game consoles, and RGB-D sensors allows human control and input, including motion-capture, to be synchronized with the visual and physics rendered environment.

\vspace{3pt}\noindent\textbf{Simulated computer vision applications.} In order to demonstrate these capabilities, we set up \simname~to generate sample synthetic data for twelve primary computer vision topics. 

In \figLabel \ref{figtab}, we present a screenshot from each of the following applications:
\textbf{(1)} object tracking;
\textbf{(2)} pose estimation;
\textbf{(3)} object detection (2D/3D);
\textbf{(4)} action recognition;
\textbf{(5)} autonomous navigation;
\textbf{(6)} 3D reconstruction;
\textbf{(7)} crowd understanding;
\textbf{(8)} urban scene understanding;
\textbf{(9)} indoor scene understanding;
\textbf{(10)} multi-agent collaboration;
\textbf{(11)} human training; and
\textbf{(12)} aerial surveying. 
In this paper, we present the full implementation and experiments on two of these CV applications: object tracking and autonomous navigation. We are releasing the full \simname~implementation of these applications, as well as, the general interface between \simname~and third party software, so as to facilitate its use in the community. In general, we believe that our simulator can provide computer vision researchers a rich environment for training and evaluating their methods across a diverse set of important applications. 

\vspace{3pt}\noindent\textbf{Unique contributions of \simname~to UE4.} Significant modifications to UE4 are made in order to enable the capabilities and applications addressed above. UE4 is re-tasked as a simulator and synthetic vision generator by the creation of new blueprints (a UE4 visual scripting language) and at a lower level bespoke C++ classes. UE4 is fully open source allowing us to exploit the full API code base to accomplish specific vision tasks that may never have been intended by UE4 creators. Unique contributions of \simname~include: 
a full Python, C++, and Matlab Socket Interface (TCP/UDP),
physics-based waypoint navigation for cars and UAVs,
PID controllers, flight and tracking controllers for UAVs,
multi-object logging and replay system,
synthetic visual data augmentation system, and an outdoor world generator with external drag-drop graphical user interface.

\begin{figure}[!htb]
\centering
\includegraphics[width = \columnwidth]{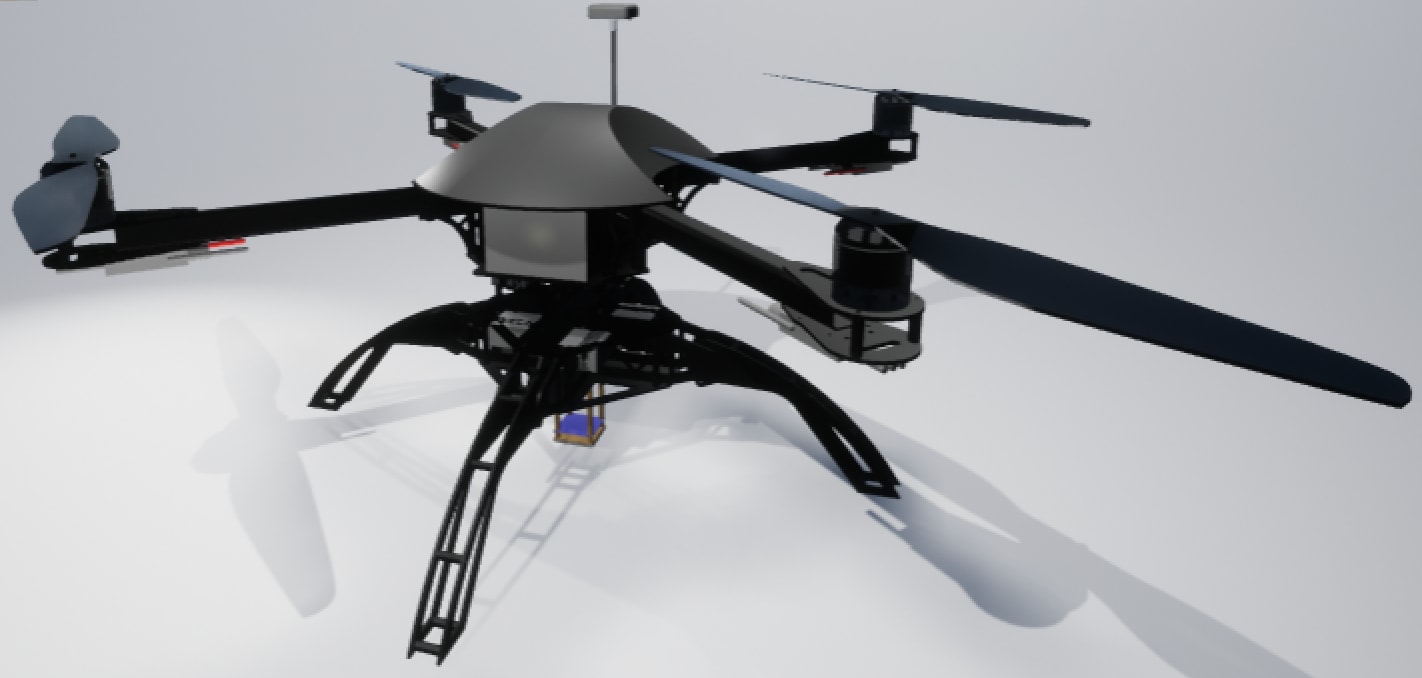}
\caption{The tracking UAV rendered inside \simname. The UAV is equipped with gimbaled landing gear allowing for both a stabilized camera and articulated legs for gripping of objects in multi-agent simulation tasks.} \label{fig:mbzirc2}
\end{figure}

UE4 provides a marketplace in which users can contribute and sell assets to the community. We purchased a broad selection of assets and modified them to work specifically with \simname. For example, the variety of cars in our simulated environment come from a set of purchased asset packs. On the other hand, the UAV used for tracking is based on a Solidworks model designed and produced by the authors to replicate a real-world UAV used to compete in a UAV challenge (refer to \figLabel \ref{fig:mbzirc2} for a rendering of this UAV).

\section{Tracking} \label{sec: abc}
\subsection{Overview}
\begin{figure}[!htb]
	\centering
	\includegraphics[width = \columnwidth]{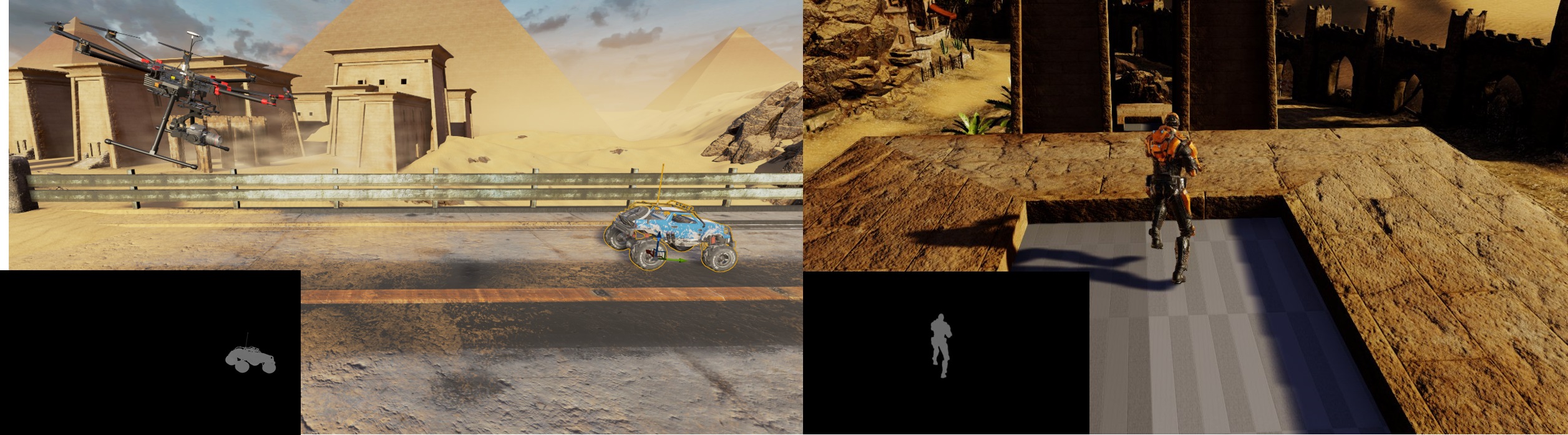}
	\caption{Two synthetic images in a desert scene generated from a virtual aerial camera within \simname~accompanied by their object-level segmentation masks.}
	\label{fig: free_gt}
\end{figure}

\begin{figure}[!htb]
	\centering
	\includegraphics[width = \columnwidth]{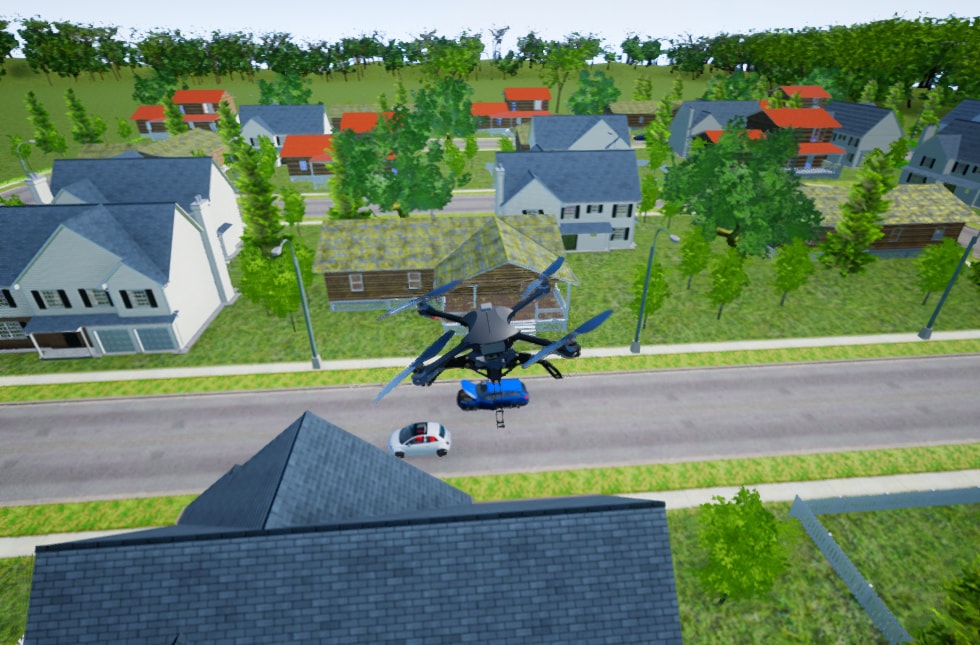}
	\caption{An image of our UAV model during online tracking evaluation.} \label{fig: tracking}
\end{figure}

One case study  application of the \simname~simulator is the ability to generate datasets with free automatic groundtruth (see \figLabel \ref{fig: free_gt}) and to evaluate state-of-the-art CV tracking algorithms "in-the-loop" under close to real-world conditions (see \figLabel \ref{fig: tracking}). In this section we demonstrate both of these capabilities. Specifically, we select five diverse state-of-the-art trackers for evaluation. These are SAMF \cite{vot14}, SRDCF \cite{srdcf}, MEEM \cite{meem}, C-COT \cite{c_cot} and MOSSE$_\text{CA}$ \cite{cf_ca_tracking}. We then perform an offline evaluation analogous to the popular tracking benchmark by \cite{28}, as is common in object tracking, but on automatically annotated sequences generated within the simulator. However, we go beyond and additionally perform an online evaluation where trackers are directly integrated with the simulator thereby controlling the UAV "on-the-fly". Preliminary results of this application were presented in \cite{Mueller2016}.

The simulator provides a test bed in which vision-based trackers can be tested on realistic high-fidelity renderings, following physics-based moving targets, and evaluated using precise ground truth annotation. Here, tracking is conducted from a UAV with the target being a moving car. As compared to the initial work in \cite{Mueller2016}, the UAV has been replaced with a new quad-copter design allowing more flight capabilities (\eg perched landing, grabbing, package delivery, \etc), and improved physics simulation using sub-stepping has been integrated. In addition, the environment and assets are now automatically spawned at game time according to a 2D map created in our new city generator. The interface with CV trackers has been redesigned and optimized allowing fast transfer of visual data and feedback through various means of communication  (\eg TCP, UDP or RAM disk). This allows easy integration of trackers written in a variety of programming languages. We have modified the chosen state-of-the-art trackers to seamlessly communicate with the simulator. Trackers that run in MATLAB can directly be evaluated within the online tracking benchmark.

\subsubsection{UAV Physics Simulation and Control}
Within UE4, the UAV is represented as a quadcopter with attached camera gimbal and gripper. A low-level flight controller maintains the UAV position and altitude within a physics based environment. Movement of the copter is updated per frame by UE4's physics simulator that accounts for Newtonian gravitational forces, input mass,  size of the UAV, and linear/angular damping. Rotating the normal of the thrust vector along the central x- and y-axis of the copter and varying thrust enables the copter to move within the environment mimicking  real-world flight. Similar to hardware-in-the-loop (HIL) approaches, the UAV control in UE4 utilizes several tuned PID controllers (written in C++ and accessed as a UE4 Blueprint function) and the calculations are done in substeps to decouple them from the frame rate.

Since we want to model a UAV in position hold, movement in the x and y directions are simulated by updating the required roll and pitch with a PID controller. The PID controller uses the difference between current and desired velocity in the x and y directions as error to be minimized. Similar to real-world tuning we experimentally adjust the controller weights within the simulator until we achieve a smooth response. Altitude of the UAV is maintained by an additional PID controller that adjusts thrust based on desired error between current altitude and desired altitude. Through this system, we are able to accurately simulate real-world flight of multi-rotors and control them by either a joystick or external input. The actual position of the UAV is kept unknown to the controller and trackers. 

\subsubsection{Extracting and Logging Flight Data}
Attached to the UAV is a camera set at a 60 degree angle and located below the frame of the copter. At every frame, the UAV camera's viewport is stored in two textures (full rendered frame and custom depth mask), which can be accessed and retrieved quickly in  MATLAB enabling the real-time evaluation of tracking within \simname. Finally, at each frame, we log the current bounding box, position, orientation, and velocity of the tracked target and the UAV for use in the evaluation of the tracker.

\subsubsection{MATLAB/C++ Integration}

The trajectory of the vehicle is replayed from a log file to ensure equal conditions for all trackers. In our experiments, trackers are setup to read frames from a RAM disk. Alternatively, trackers can communicate through TCP or UDP with the simulator. The frames are output at $320\times 180$ pixels in order to reduce latency. A script runs in MATLAB to initialize the current tracker with the initial bounding box sent by the simulator. The MATLAB script then continues to read subsequent frames and passes them to the current tracker. The output of the tracker, after processing a frame, is a bounding box, which is read by UE4 at every frame and used to calculate error between the center of the camera frame and the bounding box center. Trackers always get the most recent frame, so they drop/miss intermediate frames if their runtime is slower than the rate at which frames are being acquired. This inherently penalizes slow trackers just like in a real-world setting. 

\subsubsection{Visual Servoing}
The onboard flight control simulator updates the UAV position to bring the tracked object back to the center of the camera field-of-view. This is  done in real-time by calculating the error from the tracker's bounding box and its integration with two PID controllers. Since the camera system is mounted on a gimbal and the camera angle and altitude are held constant, only the vertical and horizontal offsets need to be calculated in the current video frame to properly reorient the UAV. The translational error in the camera frame is obtained by finding the difference between the current target's bounding box center and the center of the video frame. A fully-tuned PID controller for both the x and y dimensions receives this offset vector and calculates the proportional response of the copter movement to recenter the tracked object. The visual servoing technique employed in the simulator is robust and, with top performing trackers, it is able to follow targets across large and diverse environments. 

\subsubsection{Qualitative Tracker Performance Evaluation}
The introduction of automatic world generation allows us to fully control the environment and isolate specific tracking attributes, carry out multiple controlled experiments, and generate very diverse annotated datasets on-the-fly. Unlike real-world scenarios where the UAV and target location are not exactly known (\eg error of 5-10m due to inaccuracies of GPS), we can quantitatively compare position, orientation, and velocity of the UAV at each time-step to understand the impact of the tracker on flight dynamics.

\begin{figure}[!hb]
	\centering
	\includegraphics[width = \columnwidth]{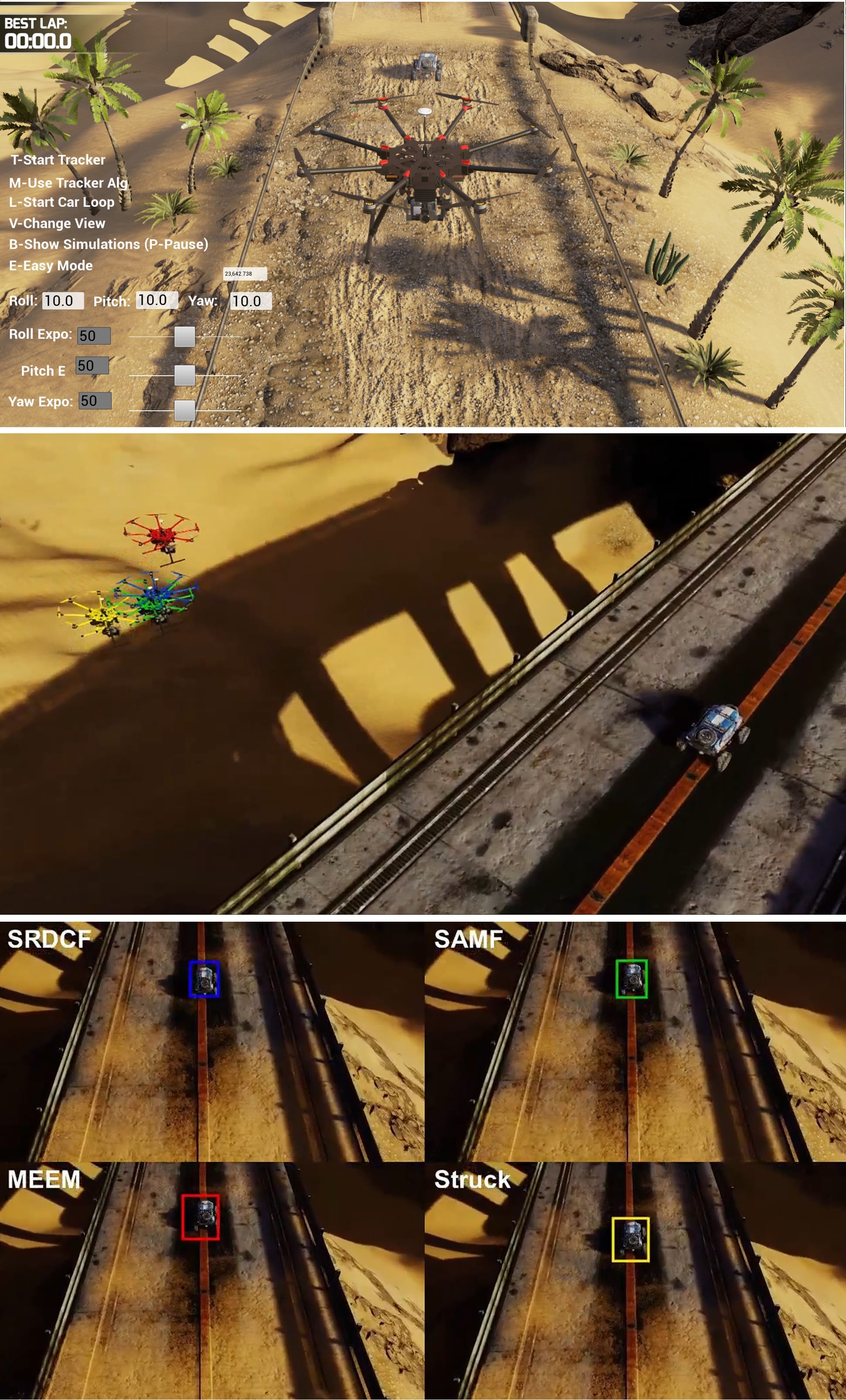} 
	\caption{\emph{Top:} Third person view of one environment in the simulator. \emph{Bottom:} Four UAVs are controlled by different trackers indicated by the different colors.} \label{fig: simulator}
\end{figure}

The simulator also enables new approaches for online performance measurement (see \figLabel \ref{fig: simulator}). For evaluation, we propose several approaches to measure tracker performance that can only be accomplished using our \simname~simulator: (1) the impact of a dynamic frame rate (different camera frame rates can be simulated and trackers are fed frames at most at the rate of computation), (2) trajectory error between the tracked target and the UAV, (3) trajectory error between a UAV controlled by ground-truth and a UAV controlled by a tracking algorithm, (4) long-term tracking within a controlled environment where attribute influence can be varied and clearly measured, and (5) recorded logs for each tracker are replayed and the path of each tracker is drawn on the 2D world map or animated in the 3D world. 

\subsection{Offline Evaluation} 
In order to demonstrate the dataset generation capabilities, we automatically generate a synthetic dataset with images captured from an UAV while following a car on the ground. The UAV uses the ground truth bounding box as input to the PID controller. We capture a total of 5 sequences from two maps with the car moving at 3 different speed levels (low: $\SI{4}{\metre\per\second}$, medium: $\SI{6}{\metre\per\second}$, high: $\SI{8}{\metre\per\second}$). The shortest sequence is $\SI{3,300}{}$ frames and the longest sequence is $\SI{12,700}{}$ frames in length. In order to benchmark tracking algorithms, we follow the classical evaluation strategy of the popular online tracking benchmark by \cite{28}. We evaluate the tracking performance using two measures: precision and success. Precision is measured as the distance between the centers of a tracker bounding box (bb\_{tr}) and the corresponding ground truth bounding box (bb\_{gt}). The precision plot shows the percentage of tracker bounding boxes within a given threshold distance in pixels of the ground truth. To rank trackers according to precision, we use the area under the curve (AUC) measure, which is also used in \cite{28}.  

Success is measured as the intersection over union of pixels in box bb\_{tr} and those in bb\_{gt}. The success plot shows the percentage of tracker bounding boxes, whose overlap score is larger than a given threshold. Similar to precision, we rank trackers according to success using the area under the curve (AUC) measure. We only perform a one-pass evaluation (OPE) in this paper.

\begin{figure}[!ht]
  \centering
  \begin{subfigure}[b]{0.999\linewidth}
    \includegraphics[width=\linewidth]{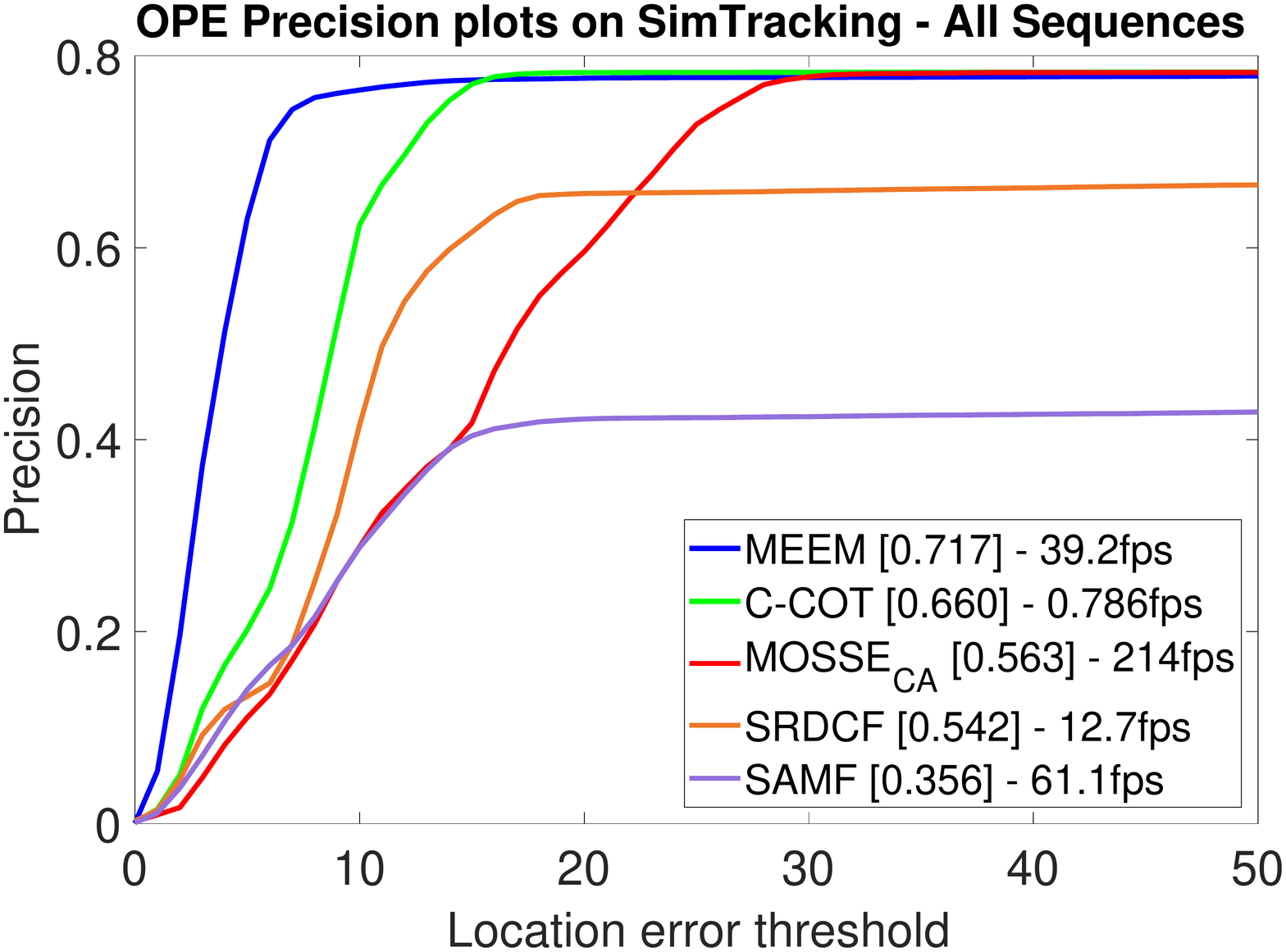}
    \label{fig:offline_precision}
  \end{subfigure}
  \vspace{-12pt}
  \begin{subfigure}[b]{0.999\linewidth}
    \includegraphics[width=\linewidth]{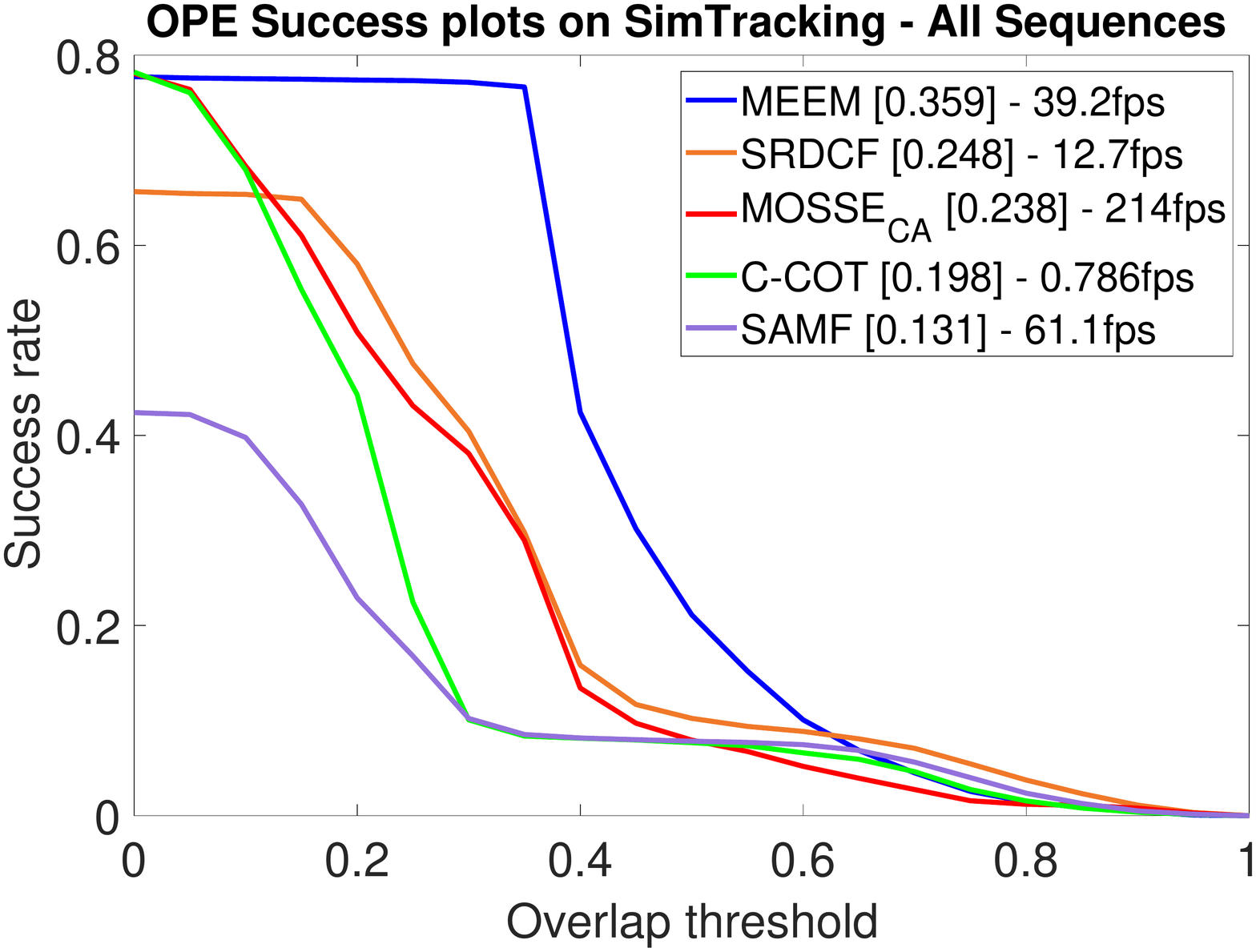}
    \label{fig:offline_success}
  \end{subfigure}
\caption{Offline evaluation: average overall performance in terms of precision and success}
\label{fig:tracking_plots}
\end{figure}

The results in \figLabel \ref{fig:tracking_plots} show that MEEM performs best in terms of both precision and success while running at over 30fps. However, note that evaluation was performed on a powerful workstation and at a low image resolution. C-COT has comparable performance but runs at a significantly lower speed (less than 1fps). Surprisingly, MOSSE$_\text{CA}$ which only used very simple features (only gray-scale pixel intensity) performs remarkably well, while running at over 200fps. We attribute this to its novel incorporation of context. Also note that at an error threshold of about 25 pixels which might still be acceptable for many applications, it is actually on par with MEEM and C-COT. SRDCF achieves similar performance as MOSSE$_\text{CA}$ and is about twenty times slower. SAMF performs significantly worse than all other trackers in this evaluation. 

\begin{figure}[!htb]
  \centering
  \includegraphics[width=\linewidth]{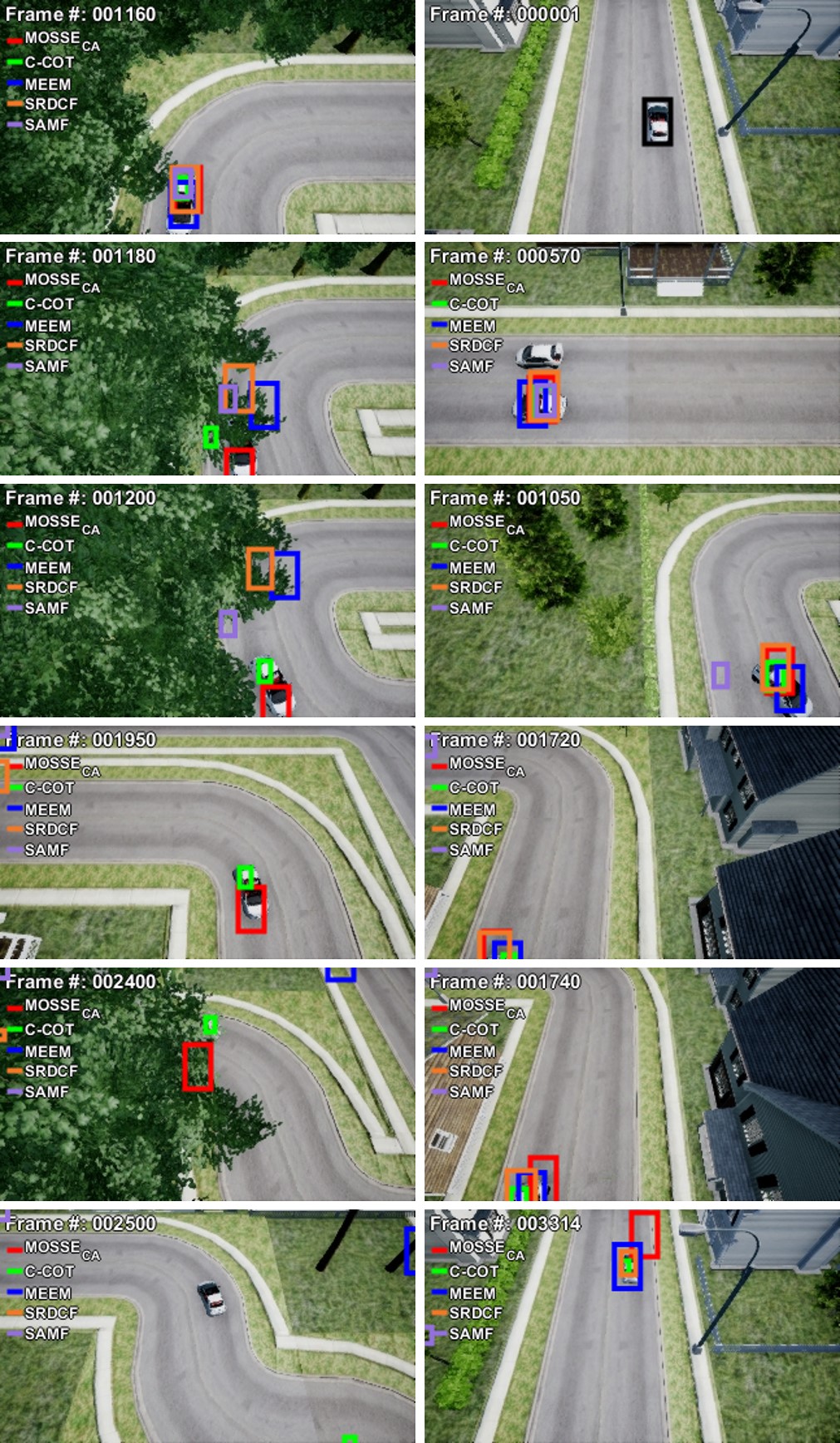}
\caption{Offline evaluation: qualitative tracking results on map1 at high speed (\emph{left column}) and map2 at medium speed (\emph{right column}). }
\label{fig:tracking_qual}
\end{figure}

\figLabel \ref{fig:tracking_qual} shows some qualitative results of the offline experiments. The images in the first row show part of a sequence in one setup, where the car is moving at high speed. SAMF already fails very early when the car drives quickly around a corner. The images in the second row show another critical moment of the same sequence, where the car goes out of view for a short period of time. Except for SAMF which already lost the target earlier, all trackers are able to track the target until the very end. Note how MOSSE$_\text{CA}$ has drifted and is only tracking a small corner of the car by the end. The third row shows a sequence of images from a much more difficult setup, where  the car is moving  at medium speed. There is heavy occlusion by a tree causing all trackers besides MOSSE$_\text{CA}$ and C-COT to lose the car. The images in the fourth row show that MOSSE$_\text{CA}$ has a much better lock on the car after this occlusion than C-COT. However, when the car gets fully occluded for several frames, no tracker is able to re-detect the car. 

\subsection{Online Evaluation}\label{simEval}

In this evaluation, the tracking algorithms communicate with the simulator. They are initialized with the first frame captured from the UAV and the corresponding ground truth bounding box. They then receive subsequent frames as input and produce a bounding box prediction as output, which is sent back to the simulator and serves as input to the PID controller of the UAV for the purpose of navigation. Depending on the speed of the tracking algorithm, frames are dropped so that only the most recent frame is evaluated much like in a real system. This evaluation emphasizes the effect of a tracker's computational efficiency on its online performance and provides insight on how suitable it would be for real-world scenarios.

We first optimize the UAV visual servoing using the ground truth (GT)  tracker, which has access to the exact position of the object from the simulator. Despite the absolute accuracy of the GT tracker, the flight mechanics of the UAV limit its ability to always keep the target centered, since it must compensate for gravity, air resistance, and inertia. After evaluating the performance of the UAV with GT, each tracker is run multiple times within the simulator with the same starting initialization bounding box.

\begin{figure*}[!htb]
	\includegraphics[width=\linewidth]{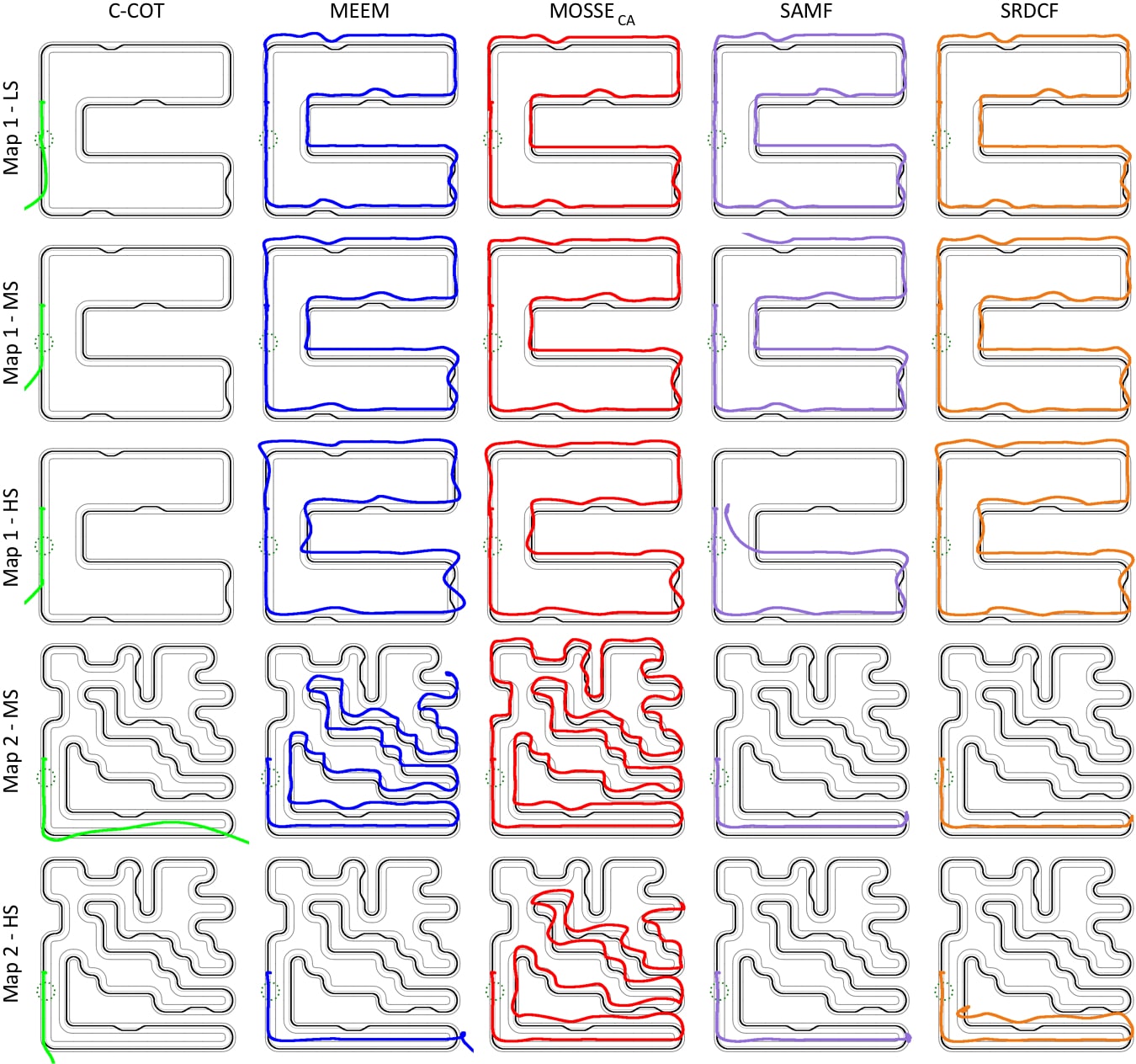} 
	\caption{Qualitative results: trajectory of 5 different trackers in 5 experiments on two different maps. LS, MS, and HS denote low, medium, and high speed characterizing the tracked object (car).}
	\label{fig:tracking_qualitative}
\end{figure*}

\figLabel \ref{fig:tracking_qualitative} shows the trajectories of the tracking algorithms on each evaluated map. SAMF is only able to complete the easy map1 when the car is moving at low speed and fails consistently otherwise. The higher the speed, the earlier the failure occurs. SRDCF, MEEM and MOSSE$_\text{CA}$ are all able to complete map1 at all speed levels. However, on map2, SRDCF fails quite early at both speeds. MEEM is able to complete about 75\% of map2 at medium speed, but fails early at high speed. MOSSE$_\text{CA}$ performs best and is the only tracker to  complete map2 at  medium speed. At  high speed, it is able to complete about 75\% of map2, again outperforming all other trackers by a margin. 

C-COT, which won the VOT16 challenge, and is currently considered the best tracking algorithm, fails to impress in this evaluation. Just to initialize and process the first frame takes about 15 seconds by which time the target is long gone. To ensure fair comparison we keep the car stationary for more than 15 seconds to provide plenty of time for initialization. However, after the car starts moving the target is lost very quickly in both maps and at all speeds, since C-COT only runs at less than 1fps resulting in very abrupt UAV motions and unstable behavior. 

\subsection{Discussion} 
MOSSE$_\text{CA}$ by \cite{cf_ca_tracking} achieves similar accuracy as the top state-of-the-art trackers, while running an order of magnitude faster. It uses a much simpler algorithm and features. This allows for deployment on a real system with limited computational resources. Further it allows processing images with larger resolution and hence more details which is especially important in the case of UAV tracking where objects are often very low resolution as showed by \cite{Mueller2016}. Lastly it permits controlling the UAV at a much faster rate if frames can be captured at higher frame rates. This also has the side effect to simplify the tracking problem since the target moves less between frames. 

\section{Autonomous Driving} \label{sec: dataset}
\subsection{Overview}
In the second case study application, we present a novel deep learning based approach towards autonomous driving in which we divide the driving task into two sub-tasks: pathway estimation and vehicle control. We train a deep neural network (DNN) to predict waypoints ahead of the car and build an algorithmic controller on top for steering and throttle control. Compared to learning the vehicle controls end-to-end, our approach has several advantages: our approach only requires auto-generated training data (waypoints), whereas a human driver would be required to generate extensive data for a supervised end-to-end approach. Additionally, our waypoint approach is more flexible, since it generalizes across different cars, even beyond the one used in training. As such,  the car can be tuned or even replaced  with a different vehicle without retraining the network. Finally, tasks such as lane changing (\secLabel \ref{sec:lanechange}), visual obstacle avoidance (\secLabel \ref{sec:obstacleavoidance}) or guided driving (\secLabel \ref{sec:guideddriving}) become quite straight-forward and easy to implement with our approach. We found that augmenting our data with respect to viewing direction is crucial for achieving high performance in these tasks. Our waypoint-based method automatically assigns real ground truth data to augmented view images as well, whereas the vehicle control outputs would have to be modified in a non-trivial way, which might require manual annotation.

To generate virtual driving environments, we develop an external software tool, where maps can be designed from a 2D overhead view, and directly imported into our simulator. From there, we automatically generate synthetic training data for our waypoint DNN. We also use the environments to evaluate our driving approach in an online fashion. This is made possible by developing an interface that enables our simulator to communicate seamlessly with the  deep learning framework (\ie TensorFlow) and thus enables our model to control the car in real-time.

\subsection{Data Acquisition}
\subsubsection{Generating Virtual Driving Environments}
\begin{figure}[!htb]
  \centering
	\begin{subfigure}[b]{0.997\linewidth}
        \includegraphics[width=\linewidth]{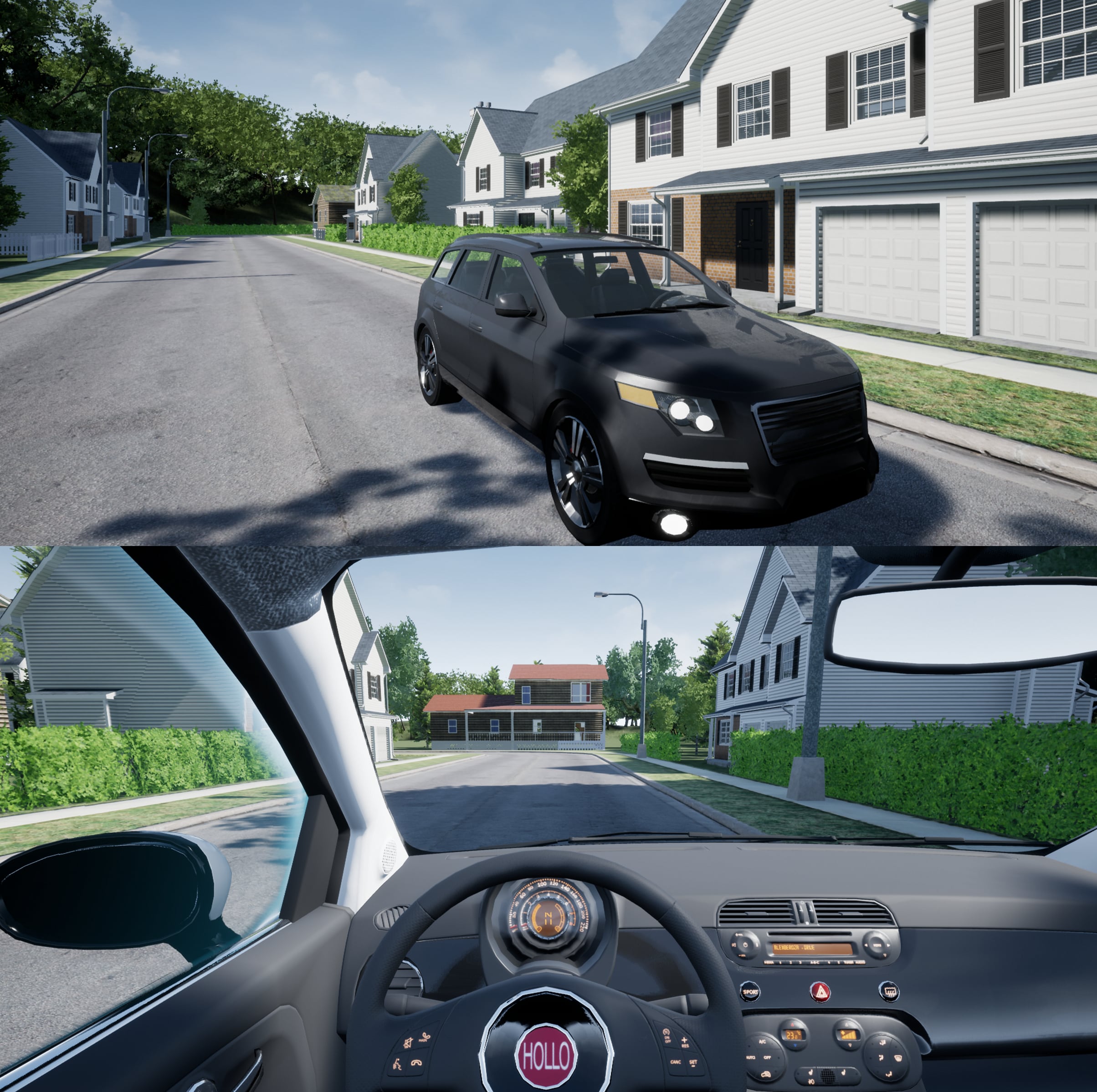}
        \caption{Images captured from our simulator in driving mode, top: third-person view, bottom: first-person view.}
        \label{fig:city_simulator}
    \end{subfigure}

	\begin{subfigure}[b]{0.997\linewidth}
        \includegraphics[width=\linewidth]{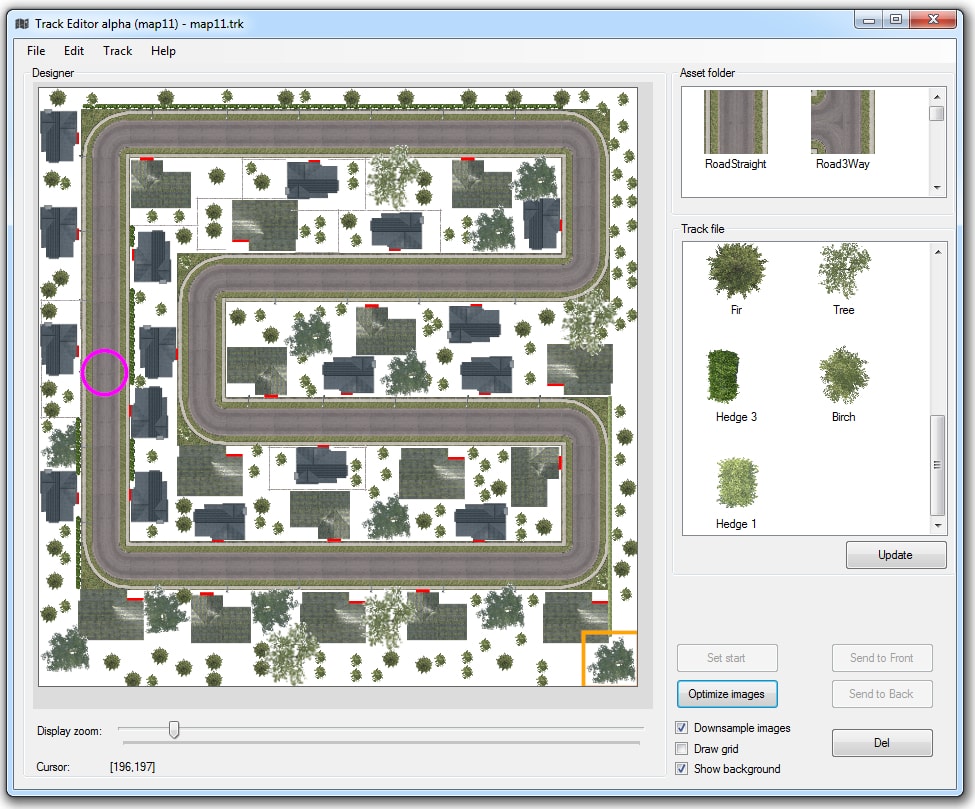}
        \caption{Our editor to create virtual driving environments. As shown in the preview, a variety of objects such as road blocks, houses or trees can be arranged on a grid.}
        \label{fig:city_editor}
    \end{subfigure}
    
    \caption{Images showing two views of a car spawned in a map, which is constructed from an overhead view using our city editor (b)}
\end{figure}

To automatically generate virtual driving environments, we develop an editor that can be used to build anything from small neighborhoods up to entire cities from an overhead view (see  \figLabel \ref{fig:city_editor}). Users can simply manipulate standardized blocks of arbitrary sizes that represent objects such as road parts, trees or houses. They can also choose to generate the road network randomly. This process can easily generate a very diverse set of training and testing environments. The editor is fully compatible with our simulator, and the generated environments can be loaded directly within \simname. \figLabel \ref{fig:city_simulator} shows images taken from our simulator while driving within an urban environment constructed using our editor in \figLabel \ref{fig:city_editor}.

\subsubsection{Generating Synthetic Image Data}
Since we are training a DNN to predict the course of the road, it is not necessary to collect human input (\ie steering angle or use of accelerator/break). Thus, to synthetically generate image data, we automatically move the car through the road system, making sure it is placed on the right lane and properly oriented, \ie at a tangent to the pathway. We render one image at every fixed distance that we move the car. However, we find that using this data alone is not sufficient to obtain a model that is able to make reasonable predictions once the car gets off the traffic lane (see \secLabel \ref{sec:augmentation}). This observation is typical for sequential decision making processes that make use of imitation or reinforcement learning. Therefore, we augment this original data by introducing two sets of parameters: x-offsets that define how far we translate the car to the left or right on the viewing axis normal, and yaw-offsets that define angles we use to rotate the car around the normal to the ground. For each original image, we investigate using fixed sets of these offsets combined exhaustively, as well as, randomly sampling views from predefined ranges of the two offset parameters.

\subsubsection{Generating Ground Truth Data}
\begin{figure}[!htb]
  \centering
  \includegraphics[width=0.995\linewidth]{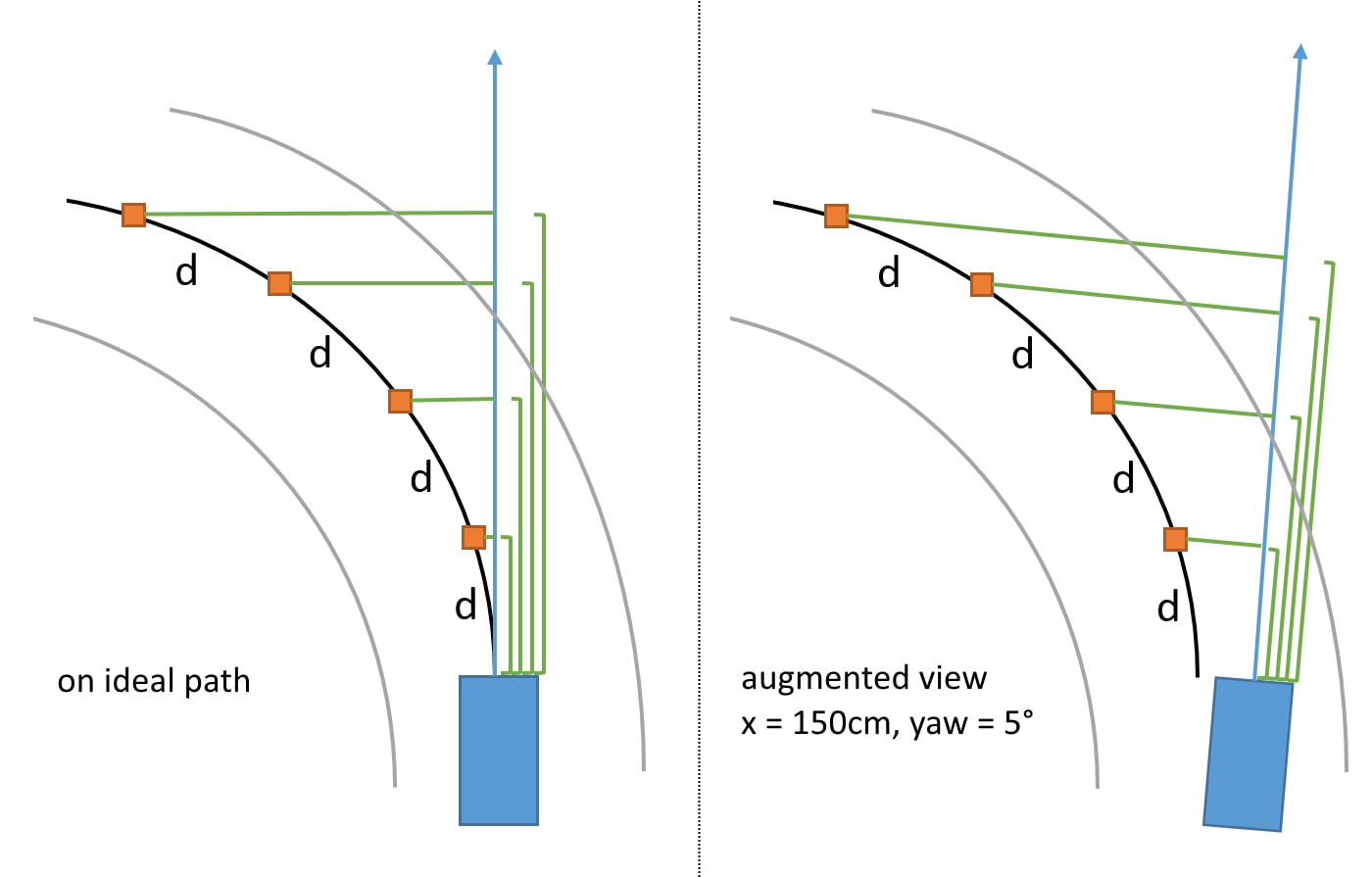}
	  \caption{An illustration of our waypoint encoding. The car (blue, viewing direction denoted by blue arrow) is about to enter a left turn. 4 waypoints (orange) are chosen such that there is a constant distance \textit{d} between them on the ideal path arc. Each waypoint is then encoded by a set of horizontal (green line segment) and vertical (green bracket) offsets.}
	\label{fig:wp_encoding}
\end{figure}

We describe the course of the road by a fixed number of waypoints that have an equal spacing. For each non-augmented view that is rendered, we choose 4 waypoints with a distance of 2 meters between each pair, so that we predict in a range of 2 to 8 meters. We then encode these 4 waypoints relative to the car position and orientation by projecting them onto the viewing axis. \figLabel \ref{fig:wp_encoding} illustrates the encoding method. We define a vertical offset that is measured as the distance between the car position and the projected point along the viewing axis (green bracket), and a horizontal offset that is defined as the distance between the original and projected point along the viewing axis normal (green line segment). For each augmentation of an original view, we  use the same four waypoints and the same encoding strategy to represent the augmented view. As such, each view (original or augmented) will be described with a rendered image and labeled with 8 \emph{ground truth} offsets using the aforementioned waypoint encoding.

We use a total of 16 driving environments (maps) for training the network (with a 12-4 split in training and validation) and an additional 4 for testing. The total road length is approximately $\SI{22,741}{\metre}$. To study the effect of context on driving performance, we setup each map in two surroundings: a sandy desert (with road only) and an urban setting (with a variety of trees, parks, houses, street lanterns, and other objects). The details of our training and test set are summarized in Table \ref{fig:data_overview}.

\begin{figure}[t]
	\begin{tabular}{ l | c | c | c}
		\hline
		 & \textbf{Training} & \textbf{Validation} & \textbf{Testing}\\ \hline
		Maps & 12 & 4 & 4 \\
        Road length & $\SI{13678}{\metre}$ & $\SI{4086}{\metre}$ & $\SI{4977}{\metre}$ \\
        Stepsize straights & $\SI{80}{\centi\metre}$ & $\SI{200}{\centi\metre}$ & - \\
        Stepsize turns & $\SI{20}{\centi\metre}$ & $\SI{200}{\centi\metre}$ & - \\
        X-offset range & $[\SI{-4}{\metre},\SI{4}{\metre}]$ & $[\SI{-4}{\metre},\SI{4}{\metre}]$ & - \\
        Yaw-offset range & $[\SI{-30}{\degree},\SI{30}{\degree}]$ & $[\SI{-30}{\degree},\SI{30}{\degree}]$ & - \\
        Random views & $\SI{1}{}$ & $\SI{3}{}$ & -\\
        Total images & $\SI{66816}{}$ & $\SI{57100}{}$ & - \\
		\hline
	\end{tabular}
	\caption{A description of our datasets and default sampling settings. We have two versions of each dataset, one in a desert and one in an urban setting, sharing the same road network.}
	\label{fig:data_overview}
\end{figure}

\subsection{DNN-Training}
\begin{figure}[!htb]
  \centering
  \includegraphics[width=0.999\columnwidth]{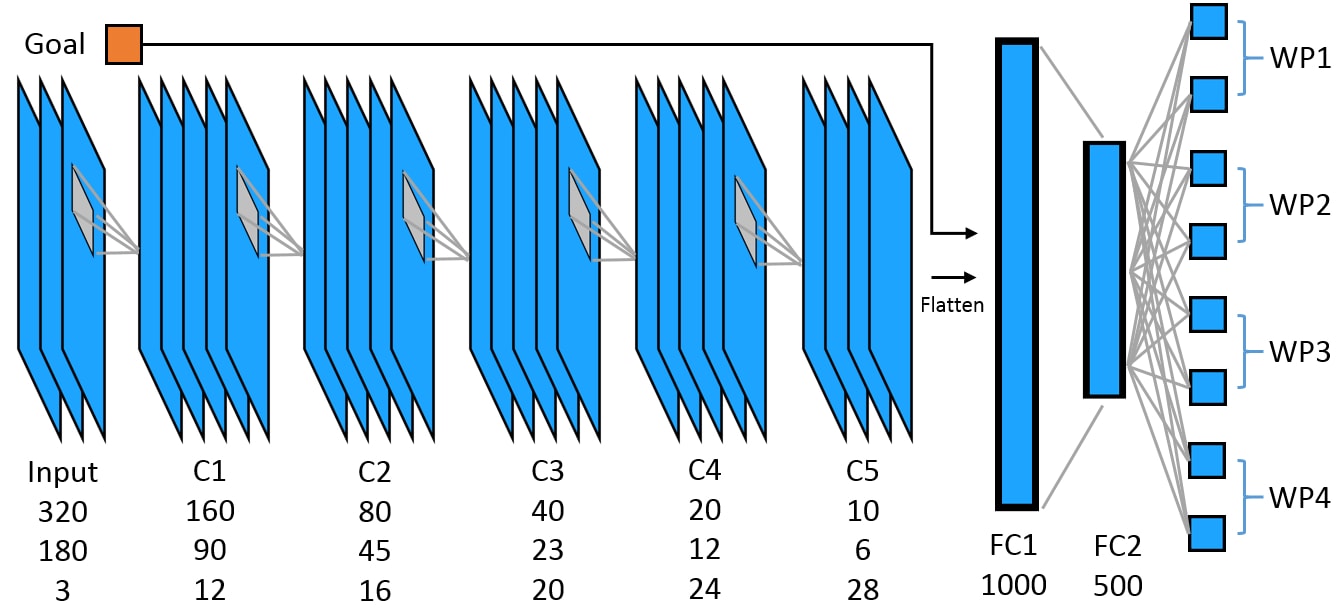}
	  \caption{Our network architecture. We use a total of 7 layers, two of which are fully connected to predict two offsets for each waypoint from a single $320\times 180$ resolution RGB-image. Optionally, we include one more value (goal) in the fully connected layer that encodes the direction the car should follow at intersections.}
	\label{fig:network_architecture}
\end{figure}

We choose the structure of our waypoint prediction network by running extensive experiments using a variety of architectures. We optimized for a small architecture that achieves high performance on the task in real-time. The structure of our best performing network is shown in \figLabel \ref{fig:network_architecture}. Notice that we optionally include one additional input (goal) that bypasses the convolutional layers. This is used to encode the desired direction at intersections for guided driving (see \secLabel \ref{sec:guideddriving}). 
The network is able to run at over 500 frames per second (fps) on an Nvidia Titan Xp when using a batch size of one (faster otherwise). We also expect it to be real-time capable on slower and/or embedded GPUs. We train our networking using a standard L2-loss and the Adam optimization algorithm, setting the base learning rate to $5\mathrm{e}{-5}$. We also use early stopping when the validation error does not decrease anymore.

\subsection{Vehicle Controller and Scoring}\label{sec:controller}
We use a simple approach to obtain steering angle and throttle from the waypoint predictions. Although there exist more sophisticated methods, we found this approach to work well in practice despite its simplicity. Based on the vertical and horizontal offset $v$ and $h$ of the first waypoint, we set the steering angle to $\theta = \arctan(\frac{h}{v})$. The throttle controller uses the horizontal offset of the last waypoint and sets the throttle based on its absolute value. For small values, throttle is high as there is a straight ahead. As the car moves closer to a turn, the absolute value increases and we lower throttle to slow down the car. Since the predictions of our network are very accurate, the trajectory that the car takes is smooth even without explicitly enforcing temporal smoothness in training or during evaluation. Of course, our approach can be easily used with different kinds of controllers that take factors like vehicle physics into account.

While there is no dedicated measurement for the controller, its performance is reflected by our measurements on the testing set. There, we run the whole system under real conditions in the simulator, whereby the car position is recorded and penalized for deviations from the  ideal pathway. We use the Euclidean distance as the penalty term and average it over the course of all tracks. For reference, the width of the car is $\SI{2}{\metre}$ and the lane width is $\SI{4}{\metre}$ each, so that the car stays exactly at the edge of the right lane on both sides if $d=\SI{1}{\metre}$. We denote the range within these bounds as the critical region. Besides measuring the average deviation, we also create cumulative histograms that denote what percentage of time the car stayed within a certain range. We use a bin size of $\SI{5}{\centi\metre}$ for the histogram.

\subsection{Evaluation}
\subsubsection{Investigating the Impact of Augmentation.}
\label{sec:augmentation}
We run a variety of experiments to investigate the impact of augmentation on our network's performance. Our results indicate that while using viewpoint augmentation is crucial, adding only a few additional views for each original frame is sufficient to achieve good performance. Without any augmentation, the system fails on every test track. Slight deviations from the lane center accumulate over time and without any augmentation, the DNN does not learn the ability to  recover from the drift. We investigate two strategies for choosing the camera parameters in augmented frames: using a set of fixed offsets at every original frame, and performing random sampling. As described in Table \ref{fig:data_overview}, we obtain random views within an x-range of $[\SI{-4}{\metre},\SI{4}{\metre}]$ and a yaw-range of $[\SI{-30}{\degree},\SI{30}{\degree}]$, providing comprehensive coverage of the road. While adding more views did not lower performance, it did not increase it either. Therefore, we choose a random sampling model that uses just one augmented view. For fixed views, we test different sets of offsets that are subsequently combined exhaustively. We find that the best configuration only utilizes rotational (yaw) offsets of $[\SI{-30}{\degree},\SI{30}{\degree}]$. Results on the test set for both augmentation strategies are shown in Table \ref{fig:augmentation}.

\begin{figure}[!htb]
  \centering
  \begin{tabular}{ l | c | c}
    		\hline
    		 & \textbf{Random views} & \textbf{Fixed views}\\ \hline
        	 \% in $[\SI{-25}{\centi \metre},\SI{25}{\centi \metre}]$ & 0.9550 & 0.8944 \\
    		 \% in $[\SI{-50}{\centi \metre},\SI{50}{\centi \metre}]$ & 0.9966 & 0.9680 \\
    		 \% in $[\SI{-1}{\metre},\SI{1}{\metre}]$ & 1.0000 & 0.9954 \\
    		 Avg deviation [\SI{}{\centi \metre}] & 7.1205 & 14.6129 \\
    		\hline
    	\end{tabular}
    	\caption{Comparison of our best performing networks trained with random sampling and fixed view augmentation, respectively. While the fixed view model still achieves very good results, it is not on par with the random sampling one.}
    	\label{fig:augmentation}
\end{figure}

We find that networks trained on fixed offsets perform worse than the ones trained on randomly augmented data. However, using the best fixed offset configuration (only yaw-offsets with $[-\SI{30}{\degree},\SI{30}{\degree}]$), we still achieve very reasonable results. This is also the setting used in several related work, including \cite{ForestTrail} and probably \cite{NvidiaCar} (two rotational offsets only are used in both papers, but the exact angles are not given in the latter). Our model is also able to navigate the car within the critical range of $[\SI{-1}{\metre},\SI{1}{\metre}]$. However, while it does outperform all human drivers, its average deviation is more than twice as high as our random view model. Our random view model stays within a very close range of $[\SI{-25}{\centi \metre},\SI{25}{\centi \metre}]$ over 95\% of the time, while it almost never leaves the range of $[\SI{-50}{\centi \metre},\SI{50}{\centi \metre}]$, compared respectively to 89\% and 97\% for the fixed view model. With an average deviation of just $\SI{7.12}{\centi \metre}$ or $\SI{14.61}{\centi \metre}$, both models drive much more accurately than our best performing human test subject at $\SI{30.17}{\centi \metre}$.

Despite training the fixed view model on 100\% more synthetic views and 50\% more data in total, it is outperformed by our random view model. We make the same observation for  models trained on even more fixed views. Since random sampling is not feasible in the real-world, this shows another advantage of using a simulator to train a model for the given driving task.

\subsubsection{Comparison to Human Performance}
\begin{figure}[!htb]
  \centering
  \includegraphics[width=0.999\linewidth]{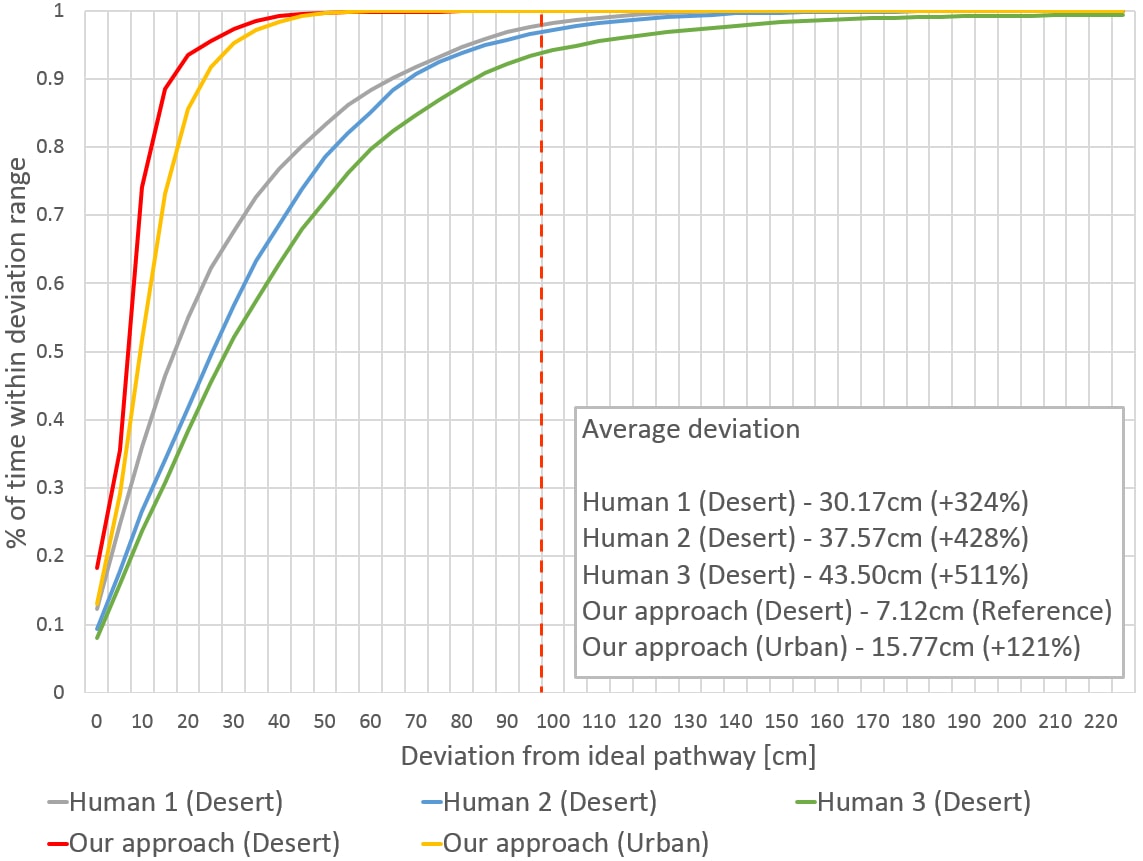}
 	  \caption{Cumulative histogram showing the deviation between ideal pathway and driven trajectory. The histogram shows the percent of time (y-axis) the car is located within the respective range (x-axis). We compare the performance of three human drivers to our system on the test tracks. Our system stays significantly closer to the ideal pathway than any human driver, and entirely avoids the critical zone (right of the dotted red line, which denotes the lane edges).}
	\label{fig:network_human}
\end{figure}

We compare the performance of our system to the driving capabilities of humans. For this, we connect a ThrustMaster Steering Wheel and Pedal Set and integrate it into our simulator. We then let three humans drive on the training maps (desert) as long as they wish. After this, we let them complete the testing maps (desert) and record results of their first and only try. We create cumulative histograms of their performance as described in \secLabel \ref{sec:controller}. We compare these results to those of our best performing network in both desert and urban environments depicted by histograms in \figLabel \ref{fig:network_human}. Clearly, all three human subjects achieve similar performance. In approximately 93-97\% of cases the human controlled car stays entirely within the lane edges ($\pm\SI{1}{\metre}$). While subjects 1 and 2 left the lane in the remaining instances, they never exceeded a distance of $\SI{140}{\centi\metre}$. This is not the case for the third test subject, who goes completely off track in a turn after losing control of the car.

In comparison, our approach is clearly superior and yields much better results. Both histograms saturate very quickly and reach 100\% at about $\SI{60}{\centi\metre}$ distance. Thus, DNN driving is much more accurate, and there is absolutely no instance where our network-driven approach navigates the car close to the lane edges. The results also suggest that our network is able to generalize to not only the given unseen test tracks, but also unseen environments not used in training, since the results on the highly altered urban environment are close to those on the desert environment.

\subsubsection{Changing Lanes}
\label{sec:lanechange}
\begin{figure}[!htb]
  \centering
  \begin{subfigure}[b]{0.999\linewidth}
    \includegraphics[width=0.999\linewidth]{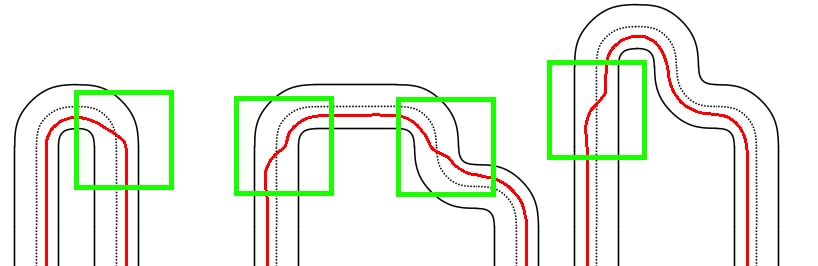}
    \subcaption{Qualitative results of our lane change procedure when called in different situations. Parts of a road map are shown from a top-down view with the driven trajectory marked in red. The lane change was initiated within the green regions.}
    \label{fig:qualitative_lanechange}
  \end{subfigure}
  
  \begin{subfigure}[b]{0.999\linewidth}
    \includegraphics[width=0.999\linewidth]{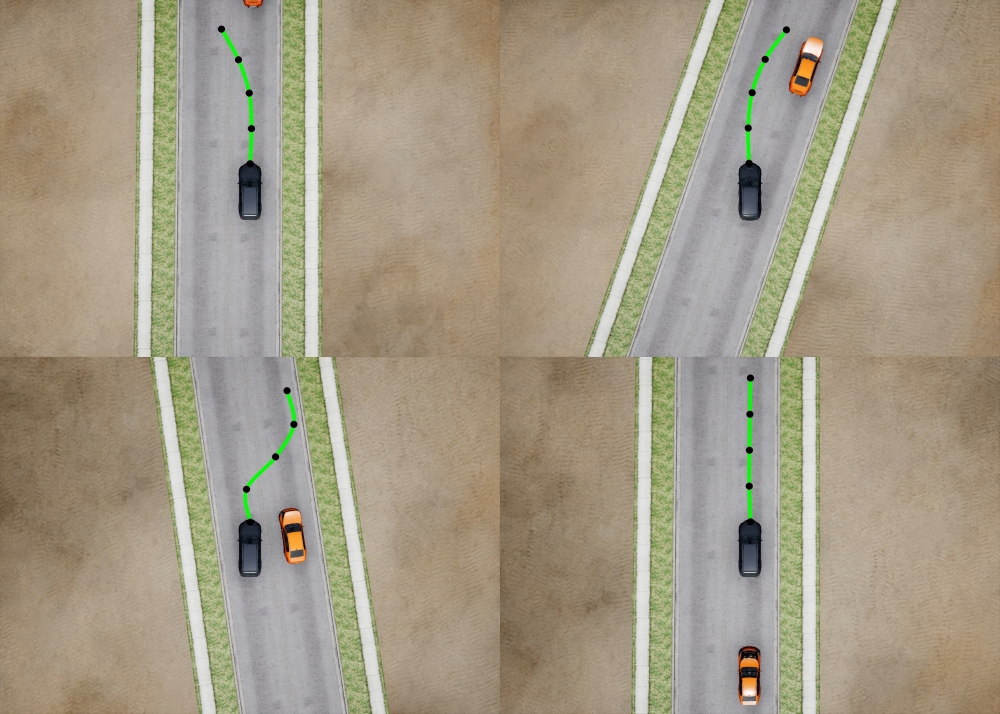}
    \subcaption{Qualitative results for visual obstacle avoidance in our desert environment. We visualized the waypoint predictions in green. The image sequence shows how the car is approaching the obstacle, performing the lane change, planning to change back to the right lane and then driving normally behind the obstacle.}
    \label{fig:qualitative_obstacleavoidance}
  \end{subfigure}
\caption{Qualitative results for lane changing and obstacle avoidance.}
\label{fig:qualitative}
\end{figure}

Changing lanes is a fundamental capability a self-driving car needs to provide. It is essential in tasks such as navigation in multi-lane scenarios or executing passing maneuvers. Our network that is trained to predict waypoints on the right lane already possesses the required capabilities to predict those on the left lane. To use them, we just flip the image before processing it in the network, and then flip the predicted horizontal offsets of each waypoint. In enabling and disabling this additional processing, the car seamlessly changes lanes accordingly. \figLabel \ref{fig:qualitative_lanechange} shows some qualitative results of our procedure. The driven trajectory is shown in red from an overhead view. Notice how the system handles a variety of cases: lane changes on straights, in ordinary turns and also S-bends. In practice, an external controller is used to trigger a lane change for this purpose. Our procedure can easily be triggered by an external controller that relies on other sensory data, as is the  case in many cars on the market today, which can determine if the desired lane is occupied. Within our simulator, it is also possible to simulate sensors, opening up interesting possibilities for more comprehensive future work in simulated environments.

\subsubsection{Learning to Avoid Obstacles}\label{sec:obstacleavoidance}
We extend our driving approach to include obstacle avoidance. For this task, we propose two approaches. One approach is to just use sensors that trigger the lane changing procedure as necessary, as discussed in \secLabel \ref{sec:lanechange}. The other (more flexible) approach is to perceive the obstacle visually and model the waypoint predictions accordingly. In this section, we present the latter, since the former has already been described.

We extend our city map editor to support obstacles that can be placed in the world as any other block element. When the ground truth (waypoints) for a map are exported, they are moved to the left lane as the car comes closer to the obstacle. After the obstacle, the waypoints are manipulated to lead back to the right lane. We extend the editor to spawn different obstacles randomly on straights and thus provide a highly diverse set of avoidance situations. After rendering these images and generating the corresponding waypoints, we follow the standard procedure for training used in obstacle-free driving described before. \figLabel \ref{fig:qualitative_obstacleavoidance} shows qualitative results in the form of an image sequence with waypoint predictions visualized in green. As it comes close to the obstacle, the car already predicts the waypoints to lead towards the other lane. It then changes back to the right lane after passing the obstacle and predicts the usual course of the road.

\subsubsection{Guided Driving}\label{sec:guideddriving}
\begin{figure}[!htb]
  \centering
  \includegraphics[width=0.999\linewidth]{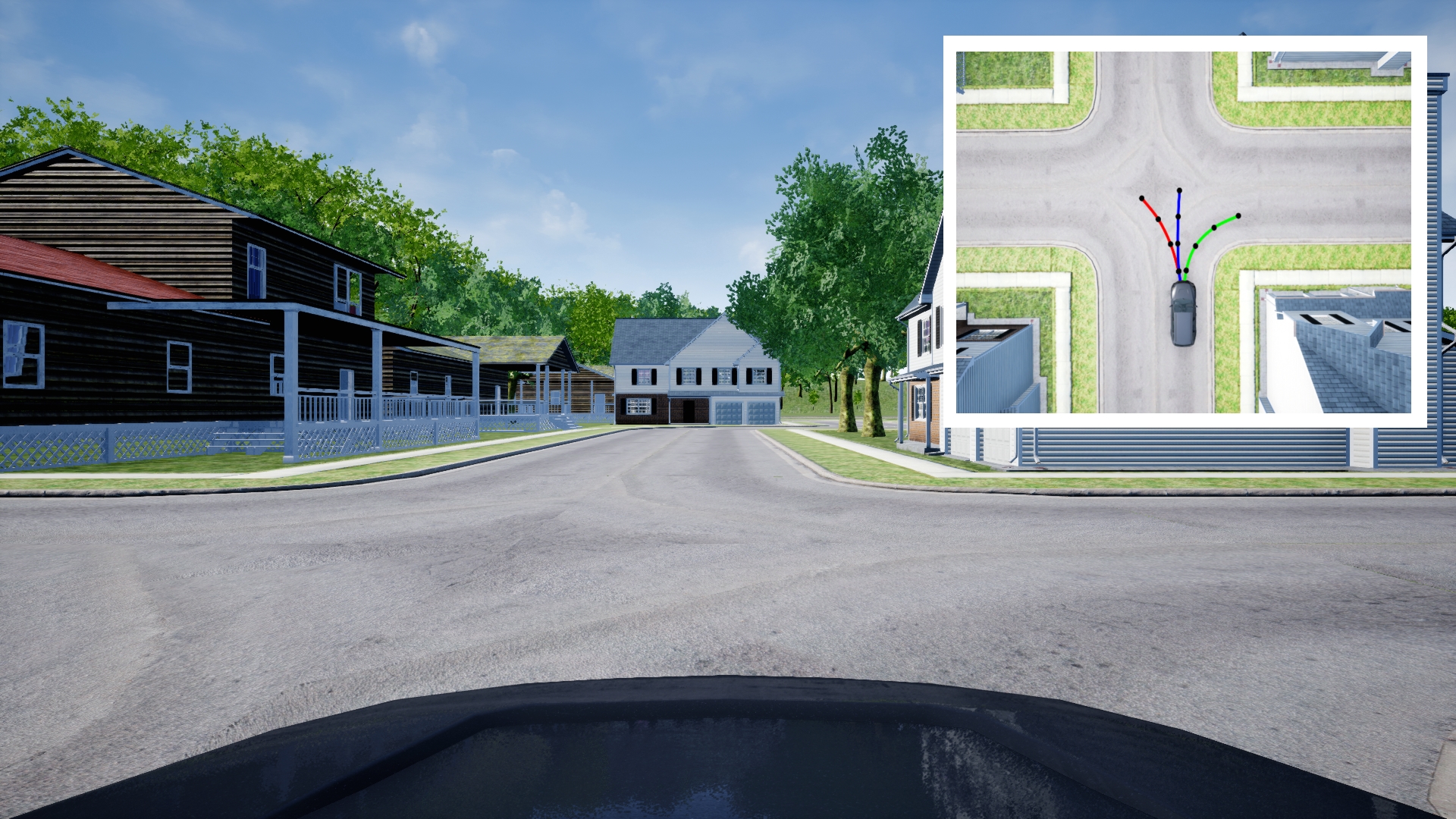} 
 	  \caption{An example showing results of our guided driving method. The car is located at an intersection. The image in the top right corner shows the situation from above with a visualization of the waypoint predictions when the goal input is set to -1 (red/left), 0 (blue/straight) and 1 (green/right), respectively.}
	\label{fig:intersection}
\end{figure}

We denote the task of controlling the car in situations of ambiguity (\ie intersections) as guided driving. This is especially useful for tasks such as GPS-based navigation, as an external controller can easily generate the appropriate sequence of directions to reach the target location. To implement this capability, we make use of the goal input in our network architecture (see \figLabel \ref{fig:network_architecture}) that feeds directly into the multilayer perceptron. We encode the intent to go left as $-1$, go straight as $0$ and go right as $+1$. We design five large maps with a variety of intersections devoted only for the guided driving task. When generating the ground truth files, we randomly decide which direction to take at intersections and make sure to fully exploit the road network of each map. We set the ground truth value of the goal input according to the direction taken if anything between the car and the furthest waypoint lies on an intersection. In other cases, we randomize the goal input as to make the network robust to changes of its value at non-intersections. We use the same configuration for training and the same random-sampling augmentation strategy. At test time, we change the goal input dynamically in the simulator. \figLabel \ref{fig:intersection} shows an example at an intersection where we parked the car and just changed the value of the goal input. The respective waypoint predictions for left, straight, and right are shown in red, blue and green, respectively. We find that learned guided driving is highly accurate in practice. In fact, in a one-hour driving test, we did not notice any mistakes.

\subsection{Discussion}
We present a novel and modular deep learning based approach towards autonomous driving. By using the deep network for pathway estimation only (thus decoupling it from the underlying car controls), we show that tasks such as lane change, obstacle avoidance, and guided driving become straightforward and very simple to implement. Furthermore, changes regarding the vehicle or its behaviour can be applied easily on the controller side without changing the learned network. Our approach even works without any need for human-generated or hand-crafted training data (although manually collected data can be included if available), thus, avoiding the high cost and tedious nature of manually collecting training data. We demonstrate the effectiveness of our approach by measuring the performance on different diversely arranged environments and maps, showing that it can outperform the capabilities of human drivers by far.

\section{Conclusions and Future Work}\label{sec: conclusion}
In this paper, we present an end-to-end high fidelity simulator that can be used for an extensive set of applications ranging across various fields of computer vision and graphics. We demonstrate the versatility of the simulator for evaluating vision problems by studying  two major applications within simulated environments, namely vision-based tracking and autonomous driving. To the best of our knowledge, this simulator is the first to provide, on both fronts, a complete real-time synthetic benchmark and evaluation system for the vision community that requires very moderate integration effort. The simulator goes beyond providing just synthetic data, but comprises a full suite of tools to evaluate and explore new environmental conditions and difficult vision tasks that are not easily controlled or replicated in the real-world.   

Although not explored in this paper, \simname~currently supports stop lights, simple walking pedestrians across crosswalks, and blueprint controlled AI cars, all of which allow for more complex driving scenarios such as navigation through crowded intersections. In future work, we hope to exploit these environmental elements of the simulator to train more sophisticated driving models. Moreover, we  pursued in this paper an SL approach to autonomous driving, but in future work we plan to pursue RL approaches for further robustness to environment variability. In this case, auxiliary sensory information can be utilized to supplement and improve vision. Of particular interest is to simulate depth measurements (simulated through stereoscopy or RGB-D/LiDAR point clouds) and provide these as new sensory data input for rewards and penalties in RL training.

In the future, we also aim to further investigate the transference of capabilities learned in simulated worlds to the real-world. One key point is to have virtual environments that are so dynamic and diverse as to accurately reflect real-world conditions. With our automated world generation system, we provide first-order functionality towards increasing graphical diversity with plans to  add more world dynamics and diversity as described above. The differences in appearance between the simulated and real-world will need to be reconciled through deep transfer learning techniques (\eg generative adversarial networks) to enable a smooth transition to the real-world. Furthermore, recent UAV work by \cite{ SadeghiL16}  highlighted that variety in rendering simulated environments enables better  transferability to the real-world, without a strong requirement on photo-realism. In the future, we hope to test a similar approach using new portable compute platforms with an attached camera in a real car or UAV.

Moreover, since our developed simulator and its seamless interface to deep learning platforms are generic in nature and open-source, we expect that this combination will open up unique opportunities for the community to develop and evaluate novel models and tracking algorithms, expand its reach to other fields of autonomous navigation, and to benefit other interesting AI tasks.


\begin{acknowledgements}
This work was supported by the King Abdullah University of Science and Technology (KAUST) Office of Sponsored Research through the VCC funding. 
\end{acknowledgements}


\bibliographystyle{spbasic}      
\bibliography{./references}

\begin{thebibliography}{61}
\providecommand{\natexlab}[1]{#1}
\providecommand{\url}[1]{{#1}}
\providecommand{\urlprefix}{URL }
\expandafter\ifx\csname urlstyle\endcsname\relax
  \providecommand{\doi}[1]{DOI~\discretionary{}{}{}#1}\else
  \providecommand{\doi}{DOI~\discretionary{}{}{}\begingroup
  \urlstyle{rm}\Url}\fi
\providecommand{\eprint}[2][]{\url{#2}}

\bibitem[{Andersson et~al(2017)Andersson, Wzorek, and Doherty}]{Andersson2017}
Andersson O, Wzorek M, Doherty P (2017) Deep learning quadcopter control via
  risk-aware active learning. In: Thirty-First AAAI Conference on Artificial
  Intelligence (AAAI), 2017, San Francisco, February 4-9., accepted.

\bibitem[{Babenko et~al(2010)Babenko, Yang, and Belongie}]{18}
Babenko B, Yang MH, Belongie S (2010) {Visual Tracking with Online Multiple
  Instance Learning.} IEEE Transactions on Pattern Analysis and Machine
  Intelligence 33(8):1619--1632, \doi{10.1109/TPAMI.2010.226}

\bibitem[{Battaglia et~al(2013)Battaglia, Hamrick, and
  Tenenbaum}]{Battaglia05112013}
Battaglia PW, Hamrick JB, Tenenbaum JB (2013) Simulation as an engine of
  physical scene understanding. Proceedings of the National Academy of Sciences
  110(45):18,327--18,332, \doi{10.1073/pnas.1306572110},
  \urlprefix\url{http://www.pnas.org/content/110/45/18327.abstract},
  \eprint{http://www.pnas.org/content/110/45/18327.full.pdf}

\bibitem[{Bojarski et~al(2016)Bojarski, Testa, Dworakowski, Firner, Flepp,
  Goyal, Jackel, Monfort, Muller, Zhang, Zhang, Zhao, and Zieba}]{NvidiaCar}
Bojarski M, Testa DD, Dworakowski D, Firner B, Flepp B, Goyal P, Jackel LD,
  Monfort M, Muller U, Zhang J, Zhang X, Zhao J, Zieba K (2016) End to end
  learning for self-driving cars. CoRR abs/1604.07316,
  \urlprefix\url{http://arxiv.org/abs/1604.07316}

\bibitem[{Brockman et~al(2016)Brockman, Cheung, Pettersson, Schneider,
  Schulman, Tang, and Zaremba}]{openaigym}
Brockman G, Cheung V, Pettersson L, Schneider J, Schulman J, Tang J, Zaremba W
  (2016) Openai gym. \eprint{arXiv:1606.01540}

\bibitem[{Chen et~al(2015)Chen, Seff, Kornhauser, and Xiao}]{deepDriving}
Chen C, Seff A, Kornhauser A, Xiao J (2015) Deepdriving: Learning affordance
  for direct perception in autonomous driving. In: Proceedings of the 2015 IEEE
  International Conference on Computer Vision (ICCV), IEEE Computer Society,
  Washington, DC, USA, ICCV '15, pp 2722--2730, \doi{10.1109/ICCV.2015.312},
  \urlprefix\url{http://dx.doi.org/10.1109/ICCV.2015.312}

\bibitem[{Collins et~al(2005)Collins, Zhou, and Teh}]{VIVID}
Collins R, Zhou X, Teh SK (2005) An open source tracking testbed and evaluation
  web site. In: IEEE International Workshop on Performance Evaluation of
  Tracking and Surveillance (PETS 2005), January 2005

\bibitem[{Danelljan et~al(2015)Danelljan, Hager, Shahbaz~Khan, and
  Felsberg}]{srdcf}
Danelljan M, Hager G, Shahbaz~Khan F, Felsberg M (2015) Learning spatially
  regularized correlation filters for visual tracking. In: The IEEE
  International Conference on Computer Vision (ICCV)

\bibitem[{Danelljan et~al(2016)Danelljan, Robinson, Shahbaz~Khan, and
  Felsberg}]{c_cot}
Danelljan M, Robinson A, Shahbaz~Khan F, Felsberg M (2016) Beyond Correlation
  Filters: Learning Continuous Convolution Operators for Visual Tracking,
  Springer International Publishing, Cham, pp 472--488.
  \doi{10.1007/978-3-319-46454-1_29},
  \urlprefix\url{http://dx.doi.org/10.1007/978-3-319-46454-1_29}

\bibitem[{De~Souza et~al(2017)De~Souza, Gaidon, Cabon, and
  Lopez~Pena}]{de2016procedural}
De~Souza C, Gaidon A, Cabon Y, Lopez~Pena A (2017) Procedural generation of
  videos to train deep action recognition networks. In: IEEE Conference on
  Computer Vision and Pattern Recognition (CVPR)

\bibitem[{Dosovitskiy et~al(2017)Dosovitskiy, Ros, Codevilla, Lopez, and
  Koltun}]{carla}
Dosovitskiy A, Ros G, Codevilla F, Lopez A, Koltun V (2017) {CARLA}: {An} open
  urban driving simulator. In: Proceedings of the 1st Annual Conference on
  Robot Learning, pp 1--16

\bibitem[{Fu et~al(2014)Fu, Carrio, Olivares-Mendez, Suarez-Fernandez, and
  Campoy}]{uavmil}
Fu C, Carrio A, Olivares-Mendez M, Suarez-Fernandez R, Campoy P (2014) Robust
  real-time vision-based aircraft tracking from unmanned aerial vehicles. In:
  Robotics and Automation (ICRA), 2014 IEEE International Conference on, pp
  5441--5446, \doi{10.1109/ICRA.2014.6907659}

\bibitem[{Furrer et~al(2016)Furrer, Burri, Achtelik, and Siegwart}]{Frrr2016}
Furrer F, Burri M, Achtelik M, Siegwart R (2016) {R}otor{S}{\textemdash}{A}
  modular gazebo {M}{A}{V} simulator framework, Studies in Computational
  Intelligence, vol 625, Springer, Cham, pp 595--625

\bibitem[{Gaidon et~al(2016)Gaidon, Wang, Cabon, and Vig}]{gaidon2016virtual}
Gaidon A, Wang Q, Cabon Y, Vig E (2016) Virtual worlds as proxy for
  multi-object tracking analysis. In: Proceedings of the IEEE Conference on
  Computer Vision and Pattern Recognition, pp 4340--4349

\bibitem[{Gaszczak et~al(2011)Gaszczak, Breckon, and Han}]{14}
Gaszczak A, Breckon TP, Han J (2011) Real-time people and vehicle detection
  from {UAV} imagery. In: R\"{o}ning J, Casasent DP, Hall EL (eds) IST/SPIE
  Electronic Imaging, International Society for Optics and Photonics, vol 7878,
  pp 78,780B--1--13, \doi{10.1117/12.876663}

\bibitem[{Ha and Liu(2014)}]{parkourSimulation}
Ha S, Liu CK (2014) Iterative training of dynamic skills inspired by human
  coaching techniques. ACM Trans Graph 34(1):1:1--1:11, \doi{10.1145/2682626},
  \urlprefix\url{http://doi.acm.org/10.1145/2682626}

\bibitem[{Hamalainen et~al(2014)Hamalainen, Eriksson, Tanskanen, Kyrki, and
  Lehtinen}]{unity3Dphysics}
Hamalainen P, Eriksson S, Tanskanen E, Kyrki V, Lehtinen J (2014) Online motion
  synthesis using sequential monte carlo. ACM Trans Graph 33(4):51:1--51:12,
  \doi{10.1145/2601097.2601218},
  \urlprefix\url{http://doi.acm.org/10.1145/2601097.2601218}

\bibitem[{Hamalainen et~al(2015)Hamalainen, Rajamaki, and
  Liu}]{balancingSimulator}
Hamalainen P, Rajamaki J, Liu CK (2015) Online control of simulated humanoids
  using particle belief propagation. ACM Trans Graph 34(4):81:1--81:13,
  \doi{10.1145/2767002}, \urlprefix\url{http://doi.acm.org/10.1145/2767002}

\bibitem[{Hejrati and Ramanan(2014)}]{syntheticCarRecognition}
Hejrati M, Ramanan D (2014) Analysis by synthesis: {3D} object recognition by
  object reconstruction. In: Computer Vision and Pattern Recognition (CVPR),
  2014 IEEE Conference on, pp 2449--2456, \doi{10.1109/CVPR.2014.314}

\bibitem[{Ju et~al(2013)Ju, Won, Lee, Choi, Noh, and
  Choi}]{birdFlightSimulator}
Ju E, Won J, Lee J, Choi B, Noh J, Choi MG (2013) Data-driven control of
  flapping flight. ACM Trans Graph 32(5):151:1--151:12,
  \doi{10.1145/2516971.2516976},
  \urlprefix\url{http://doi.acm.org/10.1145/2516971.2516976}

\bibitem[{Kendall et~al(2014)Kendall, Salvapantula, and Stol}]{Kendall2014}
Kendall A, Salvapantula N, Stol K (2014) On-board object tracking control of a
  quadcopter with monocular vision. In: Unmanned Aircraft Systems (ICUAS), 2014
  International Conference on, pp 404--411, \doi{10.1109/ICUAS.2014.6842280}

\bibitem[{Kim and Chen(2015)}]{Kim2015DeepNN}
Kim DK, Chen T (2015) Deep neural network for real-time autonomous indoor
  navigation. CoRR abs/1511.04668

\bibitem[{Koutn\'{\i}k et~al(2013)Koutn\'{\i}k, Cuccu, Schmidhuber, and
  Gomez}]{Koutnik:2013}
Koutn\'{\i}k J, Cuccu G, Schmidhuber J, Gomez F (2013) Evolving large-scale
  neural networks for vision-based reinforcement learning. In: Proceedings of
  the 15th Annual Conference on Genetic and Evolutionary Computation, ACM, New
  York, NY, USA, GECCO '13, pp 1061--1068, \doi{10.1145/2463372.2463509},
  \urlprefix\url{http://doi.acm.org/10.1145/2463372.2463509}

\bibitem[{Koutn{\'i}k et~al(2014)Koutn{\'i}k, Schmidhuber, and
  Gomez}]{Koutník2014}
Koutn{\'i}k J, Schmidhuber J, Gomez F (2014) Online Evolution of Deep
  Convolutional Network for Vision-Based Reinforcement Learning, Springer
  International Publishing, Cham, pp 260--269.
  \doi{10.1007/978-3-319-08864-8_25},
  \urlprefix\url{http://dx.doi.org/10.1007/978-3-319-08864-8_25}

\bibitem[{Kristan et~al(2014)Kristan, Pflugfelder, Leonardis, Matas,
  {\v{C}}ehovin, Nebehay, Voj{\'\i}{\v{r}}, Fernandez, Luke{\v{z}}i{\v{c}},
  Dimitriev et~al}]{vot14}
Kristan M, Pflugfelder R, Leonardis A, Matas J, {\v{C}}ehovin L, Nebehay G,
  Voj{\'\i}{\v{r}} T, Fernandez G, Luke{\v{z}}i{\v{c}} A, Dimitriev A, et~al
  (2014) The visual object tracking vot2014 challenge results. In: Computer
  Vision-ECCV 2014 Workshops, Springer, pp 191--217

\bibitem[{Lerer et~al(2016)Lerer, Gross, and Fergus}]{UE4simulator}
Lerer A, Gross S, Fergus R (2016) {L}earning {P}hysical {I}ntuition of {B}lock
  {T}owers by {E}xample. ArXiv:1603.01312v1, \eprint{1603.01312}

\bibitem[{Li et~al(2016)Li, Lin, Wu, Yang, and Yan}]{nuspro}
Li A, Lin M, Wu Y, Yang MH, Yan S (2016) {NUS-PRO}: A new visual tracking
  challenge. IEEE Transactions on Pattern Analysis and Machine Intelligence
  38(2):335--349, \doi{10.1109/TPAMI.2015.2417577}

\bibitem[{Liang et~al(2015)Liang, Blasch, and Ling}]{ColorBenchmark}
Liang P, Blasch E, Ling H (2015) Encoding color information for visual
  tracking: Algorithms and benchmark. IEEE Transactions on Image Processing
  24(12):5630--5644, \doi{10.1109/TIP.2015.2482905}

\bibitem[{Lillicrap et~al(2016)Lillicrap, Hunt, Pritzel, Heess, Erez, Tassa,
  Silver, and Wierstra}]{deepReinforcementSimulator}
Lillicrap TP, Hunt JJ, Pritzel A, Heess N, Erez T, Tassa Y, Silver D, Wierstra
  D (2016) Continuous control with deep reinforcement learning. ICLR
  abs/1509.02971, \urlprefix\url{http://arxiv.org/abs/1509.02971}

\bibitem[{Lim and Sinha(2015)}]{icraMAVtracker}
Lim H, Sinha SN (2015) Monocular localization of a moving person onboard a
  quadrotor mav. In: Robotics and Automation (ICRA), 2015 IEEE International
  Conference on, pp 2182--2189, \doi{10.1109/ICRA.2015.7139487}

\bibitem[{Mar\'{i}n et~al(2010)Mar\'{i}n, V\'{a}zquez, Ger\'{o}nimo, and
  L\'{o}pez}]{Pedestrian2010}
Mar\'{i}n J, V\'{a}zquez D, Ger\'{o}nimo D, L\'{o}pez AM (2010) Learning
  appearance in virtual scenarios for pedestrian detection. In: 2010 IEEE
  Computer Society Conference on Computer Vision and Pattern Recognition, pp
  137--144, \doi{10.1109/CVPR.2010.5540218}

\bibitem[{Mnih et~al(2016)Mnih, Badia, Mirza, Graves, Lillicrap, Harley,
  Silver, and Kavukcuoglu}]{mnih2016asynchronous}
Mnih V, Badia AP, Mirza M, Graves A, Lillicrap T, Harley T, Silver D,
  Kavukcuoglu K (2016) Asynchronous methods for deep reinforcement learning.
  In: International Conference on Machine Learning, pp 1928--1937

\bibitem[{Movshovitz-Attias et~al(2014)Movshovitz-Attias, Sheikh,
  Naresh~Boddeti, and Wei}]{syntheticVehicleTraining}
Movshovitz-Attias Y, Sheikh Y, Naresh~Boddeti V, Wei Z (2014) {3D}
  pose-by-detection of vehicles via discriminatively reduced ensembles of
  correlation filters. In: Proceedings of the British Machine Vision
  Conference, BMVA Press, \doi{http://dx.doi.org/10.5244/C.28.53}

\bibitem[{Mueller et~al(2016{\natexlab{a}})Mueller, Sharma, Smith, and
  Ghanem}]{mueller_iros16}
Mueller M, Sharma G, Smith N, Ghanem B (2016{\natexlab{a}}) Persistent aerial
  tracking system for {UAV}s. In: Intelligent Robots and Systems (IROS), 2016
  IEEE/RSJ International Conference

\bibitem[{Mueller et~al(2016{\natexlab{b}})Mueller, Smith, and
  Ghanem}]{Mueller2016}
Mueller M, Smith N, Ghanem B (2016{\natexlab{b}}) A Benchmark and Simulator for
  UAV Tracking, Springer International Publishing, Cham, pp 445--461.
  \doi{10.1007/978-3-319-46448-0_27},
  \urlprefix\url{http://dx.doi.org/10.1007/978-3-319-46448-0_27}

\bibitem[{Mueller et~al(2017)Mueller, Smith, and Ghanem}]{cf_ca_tracking}
Mueller M, Smith N, Ghanem B (2017) Context-aware correlation filter tracking.
  In: Proc. of the IEEE Conference on Computer Vision and Pattern Recognition
  (CVPR)​

\bibitem[{Muller et~al(2006)Muller, Ben, Cosatto, Flepp, and Cun}]{NIPS2005}
Muller U, Ben J, Cosatto E, Flepp B, Cun YL (2006) Off-road obstacle avoidance
  through end-to-end learning. In: Weiss Y, Sch\"{o}lkopf PB, Platt JC (eds)
  Advances in Neural Information Processing Systems 18, MIT Press, pp 739--746,
  \urlprefix\url{http://papers.nips.cc/paper/2847-off-road-obstacle-avoidance-through-end-to-end-learning.pdf}

\bibitem[{Naseer et~al(2013)Naseer, Sturm, and Cremers}]{naseer13iros}
Naseer T, Sturm J, Cremers D (2013) Followme: Person following and gesture
  recognition with a quadrocopter. In: Intelligent Robots and Systems (IROS),
  2013 IEEE/RSJ International Conference on, pp 624--630,
  \doi{10.1109/IROS.2013.6696416}

\bibitem[{Nussberger et~al(2014)Nussberger, Grabner, and
  Van~Gool}]{Nussberger2014}
Nussberger A, Grabner H, Van~Gool L (2014) Aerial object tracking from an
  airborne platform. In: Unmanned Aircraft Systems (ICUAS), 2014 International
  Conference on, pp 1284--1293, \doi{10.1109/ICUAS.2014.6842386}

\bibitem[{Papon and Schoeler(2015)}]{syntheticRGBD}
Papon J, Schoeler M (2015) Semantic pose using deep networks trained on
  synthetic {RGB-D}. CoRR abs/1508.00835,
  \urlprefix\url{http://arxiv.org/abs/1508.00835}

\bibitem[{Pepik et~al(2012)Pepik, Stark, Gehler, and Schiele}]{teaching3D}
Pepik B, Stark M, Gehler P, Schiele B (2012) Teaching {3D} geometry to
  deformable part models. In: Computer Vision and Pattern Recognition (CVPR),
  2012 IEEE Conference on, pp 3362--3369, \doi{10.1109/CVPR.2012.6248075}

\bibitem[{Pestana et~al(2013)Pestana, Sanchez-Lopez, Campoy, and
  Saripalli}]{uavtld}
Pestana J, Sanchez-Lopez J, Campoy P, Saripalli S (2013) Vision based
  {GPS}-denied object tracking and following for unmanned aerial vehicles. In:
  Safety, Security, and Rescue Robotics (SSRR), 2013 IEEE International
  Symposium on, pp 1--6, \doi{10.1109/SSRR.2013.6719359}

\bibitem[{Pollard and Antone(2012)}]{Pollard2012}
Pollard T, Antone M (2012) Detecting and tracking all moving objects in
  wide-area aerial video. In: Computer Vision and Pattern Recognition Workshops
  (CVPRW), 2012 IEEE Computer Society Conference on, pp 15--22,
  \doi{10.1109/CVPRW.2012.6239201}

\bibitem[{Pomerleau(1989)}]{pomerleau1989alvinn}
Pomerleau DA (1989) {ALVINN}: An autonomous land vehicle in a neural network.
  In: Touretzky DS (ed) Advances in Neural Information Processing Systems 1,
  Morgan-Kaufmann, pp 305--313,
  \urlprefix\url{http://papers.nips.cc/paper/95-alvinn-an-autonomous-land-vehicle-in-a-neural-network.pdf}

\bibitem[{Portmann et~al(2014)Portmann, Lynen, Chli, and
  Siegwart}]{Portmann2014}
Portmann J, Lynen S, Chli M, Siegwart R (2014) People detection and tracking
  from aerial thermal views. In: Robotics and Automation (ICRA), 2014 IEEE
  International Conference on, pp 1794--1800, \doi{10.1109/ICRA.2014.6907094}

\bibitem[{Prabowo et~al(2015)Prabowo, Trilaksono, and Triputra}]{uavHIL2015}
Prabowo YA, Trilaksono BR, Triputra FR (2015) Hardware in-the-loop simulation
  for visual servoing of fixed wing {UAV}. In: Electrical Engineering and
  Informatics (ICEEI), 2015 International Conference on, pp 247--252,
  \doi{10.1109/ICEEI.2015.7352505}

\bibitem[{Prokaj and Medioni(2014)}]{Prokaj2014}
Prokaj J, Medioni G (2014) Persistent tracking for wide area aerial
  surveillance. In: Computer Vision and Pattern Recognition (CVPR), 2014 IEEE
  Conference on, pp 1186--1193, \doi{10.1109/CVPR.2014.155}

\bibitem[{Qadir et~al(2011)Qadir, Neubert, Semke, and Schultz}]{3}
Qadir A, Neubert J, Semke W, Schultz R (2011) On-Board Visual Tracking With
  Unmanned Aircraft System (UAS), American Institute of Aeronautics and
  Astronautics, chap On-Board Visual Tracking With Unmanned Aircraft System
  (UAS). Infotech@Aerospace Conferences, \doi{10.2514/6.2011-1503}, 0

\bibitem[{Richter et~al(2016)Richter, Vineet, Roth, and Koltun}]{GtaV}
Richter SR, Vineet V, Roth S, Koltun V (2016) Playing for Data: Ground Truth
  from Computer Games, Springer International Publishing, Cham, pp 102--118.
  \doi{10.1007/978-3-319-46475-6_7},
  \urlprefix\url{http://dx.doi.org/10.1007/978-3-319-46475-6_7}

\bibitem[{Ros et~al(2016)Ros, Sellart, Materzynska, Vazquez, and
  Lopez}]{RosCVPR16}
Ros G, Sellart L, Materzynska J, Vazquez D, Lopez A (2016) {The SYNTHIA
  Dataset}: A large collection of synthetic images for semantic segmentation of
  urban scenes. In: CVPR

\bibitem[{Sadeghi and Levine(2016)}]{SadeghiL16}
Sadeghi F, Levine S (2016) {CAD2RL}: Real single-image flight without a single
  real image. CoRR abs/1611.04201,
  \urlprefix\url{http://arxiv.org/abs/1611.04201}, \eprint{1611.04201}

\bibitem[{Shah et~al(2017)Shah, Dey, Lovett, and Kapoor}]{airsim2017fsr}
Shah S, Dey D, Lovett C, Kapoor A (2017) Airsim: High-fidelity visual and
  physical simulation for autonomous vehicles. In: Field and Service Robotics,
  \urlprefix\url{https://arxiv.org/abs/1705.05065}, \eprint{arXiv:1705.05065}

\bibitem[{Shah et~al(2016)Shah, Khawad, and Krishna}]{Shah:2016}
Shah U, Khawad R, Krishna KM (2016) Deepfly: Towards complete autonomous
  navigation of {MAV}s with monocular camera. In: Proceedings of the Tenth
  Indian Conference on Computer Vision, Graphics and Image Processing, ACM, New
  York, NY, USA, ICVGIP '16, pp 59:1--59:8, \doi{10.1145/3009977.3010047},
  \urlprefix\url{http://doi.acm.org/10.1145/3009977.3010047}

\bibitem[{Smeulders et~al(2014)Smeulders, Chu, Cucchiara, Calderara, Dehghan,
  and Shah}]{VisualTrackingSurvey}
Smeulders AWM, Chu DM, Cucchiara R, Calderara S, Dehghan A, Shah M (2014)
  Visual tracking: An experimental survey. IEEE Transactions on Pattern
  Analysis and Machine Intelligence 36(7):1442--1468,
  \doi{10.1109/TPAMI.2013.230}

\bibitem[{{Smolyanskiy} et~al(2017){Smolyanskiy}, {Kamenev}, {Smith}, and
  {Birchfield}}]{ForestTrail}
{Smolyanskiy} N, {Kamenev} A, {Smith} J, {Birchfield} S (2017) {Toward
  Low-Flying Autonomous MAV Trail Navigation using Deep Neural Networks for
  Environmental Awareness}. ArXiv e-prints \eprint{1705.02550}

\bibitem[{Tan et~al(2014)Tan, Gu, Liu, and Turk}]{bikeStunts}
Tan J, Gu Y, Liu CK, Turk G (2014) Learning bicycle stunts. ACM Trans Graph
  33(4):50:1--50:12, \doi{10.1145/2601097.2601121},
  \urlprefix\url{http://doi.acm.org/10.1145/2601097.2601121}

\bibitem[{Trilaksono et~al(2011)Trilaksono, Triadhitama, Adiprawita, Wibowo,
  and Sreenatha}]{hilUAV}
Trilaksono BR, Triadhitama R, Adiprawita W, Wibowo A, Sreenatha A (2011)
  Hardware-in-the-loop simulation for visual target tracking of octorotor
  {UAV}. Aircraft Engineering and Aerospace Technology 83(6):407--419,
  \doi{10.1108/00022661111173289},
  \urlprefix\url{http://dx.doi.org/10.1108/00022661111173289}

\bibitem[{Weichao~Qiu(2017)}]{qiu2017unrealcv}
Weichao~Qiu YZSQZXTSKYWAY Fangwei~Zhong (2017) Unrealcv: Virtual worlds for
  computer vision. ACM Multimedia Open Source Software Competition

\bibitem[{Wu et~al(2013)Wu, Lim, and Yang}]{28}
Wu Y, Lim J, Yang MH (2013) {Online Object Tracking: A Benchmark}. In: 2013
  IEEE Conference on Computer Vision and Pattern Recognition, IEEE, pp
  2411--2418, \doi{10.1109/CVPR.2013.312}

\bibitem[{Wymann et~al(2014)Wymann, Dimitrakakis, Sumner, Espi\'e, Guionneau,
  and Coulom}]{torcs}
Wymann B, Dimitrakakis C, Sumner A, Espi\'e E, Guionneau C, Coulom R (2014)
  {TORCS}, the open racing car simulator. \texttt{http://www.torcs.org}

\bibitem[{Zhang et~al(2014)Zhang, Ma, and Sclaroff}]{meem}
Zhang J, Ma S, Sclaroff S (2014) {MEEM:} robust tracking via multiple experts
  using entropy minimization. In: Proc. of the European Conference on Computer
  Vision (ECCV)

\end{thebibliography}


\end{document}